\documentclass[graybox]{svmult}

\usepackage{color}
\usepackage{mathptmx}       
\usepackage{helvet}         
\usepackage{courier}        
\usepackage{type1cm}        
\usepackage{makeidx}         
\usepackage{graphicx}        
\usepackage{amsmath}                             
\usepackage{multicol}        
\usepackage[bottom]{footmisc}
\usepackage{gensymb}
\usepackage[authoryear,round,longnamesfirst]{natbib}
\makeindex             

\begin{document}

\title*{Simultaneous Localization And Mapping Without Linearization}
\author{Feng Tan, Winfried Lohmiller and Jean-Jacques Slotine}
\institute{Feng Tan \at Massachusetts Institute of Technology, \email{fengtan@mit.edu}
\and Winfried Lohmiller \at Massachusetts Institute of Technology \email{wslohmil@mit.edu}
\and Jean-Jacques Slotine \at Massachusetts Institute of Technology \email{jjs@mit.edu}}
%
%
\maketitle

\abstract{We apply a combination of linear time varying (LTV) Kalman filtering and nonlinear contraction tools to the problem of simultaneous mapping and localization (SLAM),
in a fashion which avoids linearized approximations altogether. 
By exploiting virtual synthetic measurements, the LTV Kalman observer avoids errors and approximations brought by the linearization process in the EKF SLAM. Furthermore, conditioned on the robot position, the covariances between landmarks are fully decoupled, making the algorithm easily scalable. Contraction analysis is used to establish stability of the algorithm and quantify its convergence rate. We propose four versions based on different combinations of sensor information, ranging from traditional bearing measurements and radial measurements to optical flows and time-to-contact measurements. As shown in simulations, the proposed algorithm is simple and fast, and it can solve SLAM problems in both 2D and 3D scenarios with guaranteed convergence rates in a full nonlinear context.}

\section{Introduction}
\label{sec:1}
Simultaneous localization and mapping (SLAM) is a key problem in
mobile robotics research. The Extended Kalman Filtering (EKF) SLAM
approach is the earliest and perhaps the most influential SLAM
algorithm. It is based on linearizing a nonlinear SLAM model, and then
using a Kalman filter to achieve local approximate
estimations. However, the linearization process on the originally
nonlinear model introduces accumulating errors, which can cause the
algorithm to be inconsistent and
divergent \citep{huang2007convergence, bailey2006consistency}. Such
inconsistency is particularly prominent in large-scale
problems.

This paper proposes a new approach to the SLAM problem based on
creating virtual measurements. Completely free of linearization, this
approach yields simpler algorithms and guaranteed convergence
rates. The virtual measurements also open up the possibility of exploiting
LTV Kalman-filtering and contraction analysis tools in combination.

The proposed algorithm is global and exact, which affords several
advantages over existing ones. First, the algorithm is simple and
straightforward mathematically, as it exploits purely linear kinematics
constraints. Second, following the same LTV Kalman filter framework,
the algorithm can adapt to different combinations of sensor
information in a very flexible way, and potentially extend to more
applications in navigation and machine vision and even contact based localization like \cite{dogar2010proprioceptive} and \cite{javdani2013efficient} or SLAM on jointed manipulators. Third, contraction
analysis can be easily used for convergence and consistency analysis of the
algorithm, yielding guaranteed global exponential convergence
rates. We illustrate the capability of our algorithm in providing
accurate estimations in both 2D and 3D settings by applying the proposed
framework with different combinations of sensor information, ranging
from traditional bearing measurements and range measurements to novel
ones such as optical flows and time-to-contact measurements.

Following a brief survey of existing SLAM methods in section II and of
basic contraction theory tools in Section III, the algorithm is
detailed in section IV for various combinations of sensors.
Simulation results are presented in Section V, and concluding remarks
are offered in Section VI.

\section{A Brief Survey of Existing SLAM Results}

In this section we provide a brief introduction on the problem of
simultaneous localization and mapping (SLAM). We first review the
three most popular categories of SLAM methods: extended Kalman
filter SLAM, particle SLAM and graph-based SLAM, and discuss some of
their strengths and weaknesses. We then introduce the azimuth model
that is used in this paper, along with the kinematics models describing
the locomotion of a mobile robot and the landmarks.

Simultaneous localization and mapping is one of key problems in mobile robotics research. SLAM is concerned about accomplishing two tasks simultaneously: mapping an unknown environment with one or multiple mobile robots and localizing the mobile robot/robots. One common model of the environment consists of multiple landmarks such as objects, corners, visual features, salient points, etc. represented by points. And a coordinate vector is used to describe the location of each landmark in 2D or 3D space.  

There are three main categories of methods for SLAM: EKF SLAM,
graph-based SLAM, and particle filter SLAM. EKF
SLAM \citep{moutarlier1990experimental, cheeseman1987stochastic, smith1990estimating, 1981363}
uses the extended Kalman
filter \citep{jazwinski2007stochastic, kalman1960new}, which
linearizes and approximates the originally nonlinear problem using the
Jacobian of the model to get the system state vector and covariance
matrix to be estimated and updated based on the environment
measurements.

Graph-based SLAM \citep{konolige2004large,montemerlo2006large,grisetti2010tutorial,folkesson2004graphical,duckett2002fast,dellaert2006square,thrun2006graph} uses graph relationships to model the constraints on estimated states and then uses nonlinear optimization methods to solve the problem. The SLAM problem is modeled as a sparse graph, where the nodes represent the landmarks and each instant pose state, and edge or soft constraint between the nodes corresponds to either a motion or a measurement event. Based on high efficiency optimization methods that are mainly offline and the sparsity of the graph, graphical SLAM methods have the ability to scale to deal with much larger-scale maps.

The particle method for SLAM relies on particle filters \citep{matthies1987error}, which enables easy representation for multimodal distributions since it is a non-parametric representation. The method uses particles representing guesses of true values of the states to approximate the posterior distributions. The first application of such method is introduced in \cite{doucet2000rao}. The FastSLAM introduced in \cite{montemerlo2002fastslam} and \cite{montemerlo2007fastslam} is one of the most important and famous particle filter SLAM methods. There are also other particle filter SLAM methods such as \cite{del1999central}.

However, each of these three methods has weaknesses and limitations: 

For the EKF SLAM, the size of the system covariance matrix grows quadratically with the number of features or landmarks, thus heavy computation needs to be carried out in dense landmark environment. Such issue makes it unsuitable for processing large maps. Also, since the linearized Jacobian is formulated using estimated states, it can cause inconsistency and divergence of the algorithm \citep{huang2007convergence, bailey2006consistency}.

For graph-based SLAM, because performing the advanced optimization methods can be expensive, they are mostly not online. Moreover, the initialization can have a strong impact on the result.

Lastly, for particle methods in SLAM, a rigorous evaluation in the number of particles required is lacking; the number is often set manually relying on experience or trial and error. Second, the number of particles required increases exponentially with the dimension of the state space. Third, nested loops and extensive re-visits can lead to particles depletion, and make the algorithm fail to achieve a consistent map.

Our method generally falls into the category of Kalman filtering SLAM. By exploiting contraction analysis tools and virtual measurements, our algorithm in effect builds a stable linear time varying(LTV) Kalman filter. Therefore, compared to the EKF SLAM methods, we do not suffer from errors brought by linearization process, and long term consistency is guaranteed. The math is simple and fast, as we do not need to calculate any Jacobian of the model. And the result we achieve is global, exact and contracting in an exponential favor. 
\\

\noindent\textbf{The Azimuth Model of the SLAM Problem in Local Coordinates}

\begin{figure*}
\includegraphics[width=0.6\textwidth]{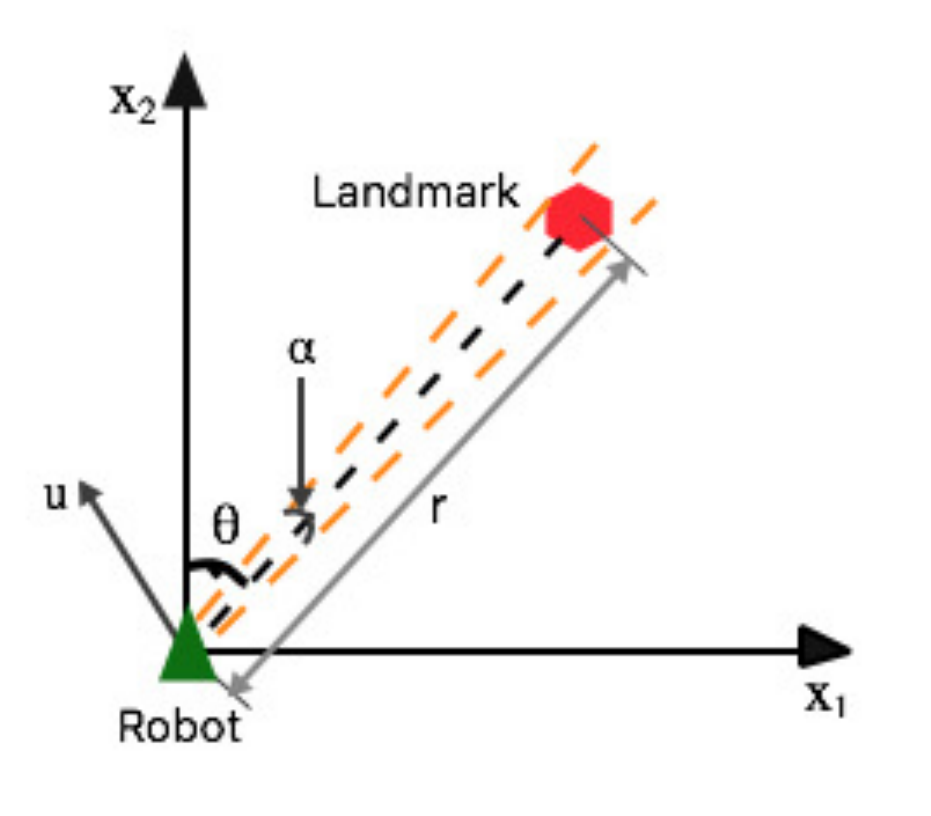}
\centering
\caption{Various types of measurements in the azimuth model}
\label{azimuth}
\end{figure*}

Let us first introduce the linear model of SLAM in local coordinates, where we use the azimuth model which measures the azimuth angle in an inertial reference coordinate $C_l$ fixed to the center of the robot and rotates with the robot (Fig.\ref{azimuth}), as in \cite{lohmiller2013contraction}. The robot is a point of mass with position and attitude.

The actual location of a landmark is described as ${\bf x}=(x_1,
x_2)^T$ for 2D and $(x_1, x_2, x_3)^T$ for 3D. The measured azimuth
angle from the robot is $\theta= \arctan (\frac{x_1}{x_2})$. In 3D
there is also the pitch measurement to the landmark $\phi=\arctan
(\frac{x_3}{\sqrt{x_1^2+x_2^2}})$. The robot's translational velocity is
${\bf u} =(u_1, u_2)^T$ in 2D, and $(u_1, u_2, u_3)^T$ in
3D. $\Omega$ is the angular velocity matrix of the robot: in the 2D
case, ${\Omega}=\begin{bmatrix} 0& -\omega_z \\ \omega_z& 0
\end{bmatrix}$ and in the 3D case, ${\Omega}=\begin{bmatrix}
0& -\omega_z&\omega_y \\
\omega_z&0& -\omega_x\\
 -\omega_y&\omega_x&0\\
\end{bmatrix}$. In both cases the matrix ${\Omega}$ is skew-symmetric.\\\\

For any landmark ${\bf x}_i$ in the inertial coordinate fixed to the robot, the relative motion is: $$\dot{{\bf x}}_i = -\Omega {\bf x}_i - {\bf u}$$
where  both $\bf u$ and $\Omega$ are assumed to be measured accurately, a reasonable assumption in most applications.

If available, the range measurement from the robot to the landmark is \\
$r = \sqrt{x_1^2+x_2^2} \ $ in 2D, and $r = \sqrt{x_1^2+x_2^2+x_3^2} \ $ in
3D.

\begin{figure*}
\includegraphics[width=0.6\textwidth]{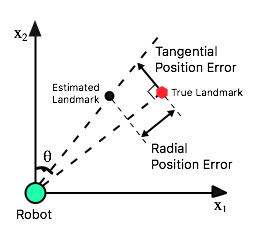}
\centering
\caption{Position error between estimation and true landmark in Cartesian coordinates}
\label{azimuthnew}
\end{figure*}

\section{Basic Tools in Contraction Theory}
\label{sec:2}
Contraction theory \citep{lohmiller1998contraction} is a relatively recent dynamic analysis and design tool, which is an exact differential analysis of convergence of complex systems based on the knowledge of the system's linearization (Jacobian) at all points. Contraction theory converts a nonlinear stability problem into an LTV (linear time-varying) first-order stability problem by considering the convergence behavior of neighboring trajectories. While Lyapunov theory may be viewed as a ``virtual mechanics'' approach to stability analysis, contraction is motivated by a ``virtual fluids'' point of view. Historically, basic convergence results on contracting systems can be traced back to the numerical analysis literature \citep{lewis1949metric, Hartman,Demidovich}.

\noindent{\bf{Theorem in \cite{lohmiller1998contraction}:}}
{\itshape
Given the system equations $\dot{\bf x} = {\bf f(x},t)$, where ${\bf f}$ is a differentiable nonlinear complex function of ${\bf x}$ within $C^n$.  If there exists a uniformly positive definite metric 
${\bf M}$ such that 
$$ {\bf \dot{M}+M\frac{\partial f}{\partial x}+\frac{\partial f}{\partial x}}^T {\bf M} \le -\beta_{M} {\bf M}$$
with constant $\beta_M>0$, then all system trajectories converge exponentially to a
single trajectory, which means contracting, with convergence rate $\beta_M$}.

Depending on the application, the metric can be found trivially (identity or rescaling of states), or obtained from physics (say, based on the inertia tensor in a mechanical system as e.g. in \cite{lohmiller2008shaping,lohmiller2013exact}). The reader is referred to \cite{lohmiller1998contraction} for a discussion of basic features in contraction theory.

\section{Landmark Navigation and LTV Kalman Filter SLAM} \label{landmark}

In this section we illustrate the use of both LTV Kalman filter and contraction tools on the
problem of navigation with visual measurements, an application often
referred to as the landmark (or lighthouse) problem, and a
key component of simultaneous localization and mapping (SLAM).

The main issues for EKF SLAM lie in the linearization and the inconsistency caused by the approximation. Our approach to solve the SLAM problem in general follows the paradigms of LTV Kalman filter. And contraction analysis adds to the global and exact solution with stability assurance because of the exponential convergence rate. 

We present the results of an exact LTV Kalman observer based on the
Riccati dynamics, which describes the Hessian of a Hamiltonian
p.d.e. \citep{lohmiller2013contraction}. A rotation term similar to
that of \cite{6846293} in the context of perspective vision systems
is also included.

\subsection{LTV Kalman filter SLAM using virtual measurements in local coordinates}

A standard extended Kalman Filter design \citep{bryson1975applied}
would start with the available nonlinear measurements, for example in
2D (Fig.\ref{azimuth})

$$\theta =\arctan \big(\frac{x_1}{x_2}\big) \ \ \ \ \ \ \ \text{and/or} \ \ \ \ \ \ \ r = \sqrt{x_1^2+x_2^2}$$

\noindent and then linearizes these measurements using the estimated
Jacobian, leading to a locally stable observer. Intuitively, the starting point of
our algorithm is the simple remark that the above relations can be
equivalently written in Cartesian coordinates as

$$ {\bf hx}=0 \ \ \ \ \ \ \ \ \ \text{and/or} \ \ \ \ \ \ \ \ \ {\bf h^*x}=r$$

\noindent where for 2D scenarios

$${\bf h}=(\cos\theta, -\sin\theta)  \ \ \ \ \ \ \ \ \ \ \ \ \ \ {\bf h^*}=(\sin\theta, \cos\theta)$$

\noindent and for 3D scenarios:
$$
{\bf h} = \left( \begin{array}{ccc} \cos \theta & -\sin \theta & 0 \\ 
-\sin \phi \sin \theta & -\sin \phi \cos \theta & \cos \phi \end{array} \right)\
$$ 
$$
{\bf h^*}=(\cos\phi \sin\theta, \cos\phi \cos\theta, \sin\phi)\
$$

\noindent Indeed in the azimuth model, one has in 2D
$${\bf x}=(x_1, x_2)^T=(r\sin{\theta}, r\cos{\theta})^T$$
and in 3D
$${\bf x}=(x_1, x_2)^T=(r\cos{\phi}\sin{\theta}, r\cos{\phi}\cos{\theta}, r\sin{\phi})^T$$

\noindent Thus, instead of directly comparing the measurement and the observation of $\theta$ and $r$,  we choose to have feedbacks on the tangential and radial Cartesian position errors between estimated and true landmark positions using simple geometrical transformation. So instead of the nonlinear observations, we can have the linear substitutes using virtual measurements and further exploit these exact linear time-varying expressions to achieve a globally stable observer design. This simple philosophy can be easily extended to a variety of SLAM contexts with different measurement inputs, especially in the visual SLAM field.

All our propositions in the following cases have the same continuous LTV Kalman filter structure using the
implicit measurements $${\bf y} = {\bf H} {\bf x} + {\bf v}(t) $$ where $\bf y$ is the measurement/observation vector in Cartesian coordinates, which includes both actual and virtual measurements; $\bf H$ is the observation model matrix, consisted of state-independent measurement vectors such as ${\bf h}$ and ${\bf h^*}$; ${\bf v}(t)$ is a zero-mean white noise also in Cartesian coordinates with the covariance ${\bf R}$.

The filter consists of two differential equations, one for the state estimations and one for the covariance updates:
$$
\dot{\hat{{\bf x}}}=-{\bf u}-{\Omega}\bf{\hat{x}}+K(y-H\hat{x})$$
$$
\dot{{\bf P}}={\bf Q-PH}^T{\bf R}^{-1}{\bf HP}-\Omega{\bf P+P}\Omega$$
with the Kalman gain ${\bf K}$  given by
$$
{\bf K}={\bf PH}^T{\bf R}^{-1}
$$
and
$${\bf y}={\bf Hx}+ {\bf v}(t)$$
$${\bf Q}=cov({\bf w})$$

\noindent where ${\bf w}(t)$ is a zero-mean white noise included in $\bf u$ (due e.g. to motion measurement inaccuracy, or rough or slippery terrain). 

The definition of $\bf y$ and $\bf H$  varies according to the type of measurement available, as we now discuss.
\\
\\
\noindent\textbf{Case I: bearing measurement only}

This original version of bearing-only SLAM was presented
in \cite{lohmiller2013contraction} where:
$$y=0 \quad\quad\quad\quad{\bf H}= 
{\bf h} $$
Geometrically, the
  virtual measurement error term $\bf h\hat{x}$ corresponds to rewriting an
  angular error as a tangential position error between estimated and
  true landmark positions. As the vehicle moves, at any instant the system contracts exponentially in the tangential direction if ${\bf
    R}^{-1}>0$, and it is indifferent along the unmeasured
  radial direction.\\

\noindent\textbf{Case II: bearing with range measurement}

If we have both bearing measurement $\theta$  and $\phi$ and range measurement $r$, a new constraint would be $y_2=r={\bf h^*x}$. 

So that for both 2D and 3D the virtual measurement is:
$${\bf y}=\begin{bmatrix}
{y_1} \\
\ {y_2} 
\end{bmatrix}=\begin{bmatrix}
{0} \\
r
\end{bmatrix} $$$${\bf H}= 
\begin{bmatrix}
{\bf h} \\
\ {\bf h*} 
\end{bmatrix} $$

\noindent\textbf{Case III: bearing with independent $\dot\theta$ information}
\\
In this case we utilize $\dot\theta$ as additional information. $\dot\theta$ is the measured relative angular velocity from the robot to the landmark and we also have $\dot\phi$ in the 3D case. Independent $\dot\theta$ measurement could be achieved either computationally based on $\theta$ or through optical flow algorithms on visual sensors. We propose here that $\dot\theta$ gives us an additional dimension of information that helps the LTV Kalman filter with radial contraction. The additional constraint or observation we get is based on the relationship  $$range \times angular\ velocity = tangential\ velocity$$ where in our case $\bf h^*\hat{x}$ is the length of vector $\bf \hat{x}$ projected along azimuth direction to represent the estimated range. 

So if the estimation is precise, $\dot{\theta}{\bf h^*\hat{x}+h}\Omega {\bf\hat{x}}$  should equal to $\bf -hu$, which is the relative velocity projected along the tangential direction.

In this case, $y_1={\bf hx}=0$ is the constraint on bearing measurement and ${y_3}=(\dot{\theta}{\bf h^*+h}\Omega){\bf x =-hu}$ in 2D or ${\bf  y_3}= (\begin{bmatrix}
\dot{\theta}\begin{bmatrix}\sin{\theta}&\cos{\theta}&0\end{bmatrix}\\
\dot{\phi}{\bf h^*}
\end{bmatrix}+ {\bf h}\Omega) {\bf x =-hu}$ in 3D is the constraint about relative angular velocity, radial distance and the tangential velocity. 

The same constraint could be derived similarly from:
$$\frac{d}{dt}{\bf hx}=0 \Rightarrow \dot{\bf h}{\bf x}+{\bf h}\dot{\bf x}=0$$
which means $$\dot{\bf h}{\bf x}+{\bf h}(\Omega {\bf x+u})=0$$
where $\dot{\bf h}=\dot{\theta}{\bf h^*}$ in 2D and $\dot{\bf h}=\begin{bmatrix}
\dot{\theta}\begin{bmatrix}\sin{\theta}&\cos{\theta}&0\end{bmatrix}\\
\dot{\phi}{\bf h^*}
\end{bmatrix}$ in 3D. Such derivation achieves the same result as the constraint we proposed earlier about tangential velocity. So the virtual measurement consists of two parts:
$${\bf y}=\begin{bmatrix}
{y_1} \\
{y_3} 
\end{bmatrix}=\begin{bmatrix}
{0} \\
-{\bf hu}
\end{bmatrix}$$
with the observation model: (2D)
$H=
\begin{bmatrix}
{\bf h} \\
\ \dot{\theta}{\bf h^*+h}\Omega
\end{bmatrix}$ \ \ \ \ \ \ \
	(3D) $H=
\begin{bmatrix}
{\bf h} \\
\ \begin{bmatrix}
\dot{\theta}\begin{bmatrix}\sin{\theta}&\cos{\theta}&0\end{bmatrix}\\
\dot{\phi}{\bf h^*}
\end{bmatrix}+ {\bf h}\Omega
\end{bmatrix}$
\\
\\
\noindent\textbf{Case IV: bearing with time to contact measurement $\tau$}

In this case we utilize the ``time to contact" measurement as additional information. Time-to-contact \citep{horn2007time,Browning2012,clady} measurement provides an estimation of time to reach the landmark, which could suggest the radial distance to the landmark based on local velocity information. This is one popular measurement for sailing and also utilized by animals and insects. For a robot, the ``time to contact" measurement could be potentially achieved by optical flows algorithms, direct gradient based methods like \cite{horn2007time}, or some novel sensors specifically developed for that purpose. 
$$
\tau=|\frac{\alpha}{\dot\alpha}| \approx |\frac{r}{\dot r}| 
$$
As shown in Fig.\ref{azimuth}, we can get the measurement $\tau=|\frac{\alpha}{\dot\alpha}|$, where $\alpha$ is an small angle measured between two feature points, edges on a single distant landmark for example. In our case, we use the angle between two edges of the cylinder landmark so that $\alpha\approx arctan(\frac{d}{r})$, where $d$ is the diameter of the cylinder landmark and $r$ is the distance from the robot to the landmark. Thus in this case besides the bearing constraint $y_1={\bf hx}=0$, we propose a novel constraint $y_4$ utilizing the ``time to contact" $\tau$.

As we know $r= {\bf h^*x}$ so that $$\dot{r}={\bf -h^*u-h}^*{\Omega}{\bf{x}+\dot{h}^*x}$$
Since $\bf h^*$ is the unit vector with the same direction of $\bf x$, both ${\bf h}^*\Omega {\bf x}$ and $\bf \dot{h}^*x$ equal to 0, so simply $\dot{r}={\bf -h^*u}$, and $$\tau=|\frac{r}{\dot r}|=\frac{{\bf h^*x}}{|{\bf -h^*u}|}$$

\noindent which means: $|\tau {\bf h^*u}| \approx {\bf h^*x}$
so we can have ${y_4}=|\tau {\bf h^*u}|$

$${\bf y}=\begin{bmatrix}
{y_1} \\
{y_4} 
\end{bmatrix}=\begin{bmatrix}
{0} \\
|\tau {\bf h^*u}|
\end{bmatrix} $$$${\bf H}= 
\begin{bmatrix}
{\bf h} \\
\ {\bf h^*} 
\end{bmatrix}$$

So that ${\bf y=Hx+v}$, and it is applicable to both 2D and 3D cases. One thing to notice is that the time-to-contact measurement is an approximation. Also when ${\bf uh}\approx 0$, $\tau$ would be reaching infinity, which reduces the reliability of the algorithm near that region.
 \\\\
\noindent\textbf{Case V: range measurement only}

If the robot has no bearing information, it may still perform SLAM if range measurements 
and their time-derivatives are available. Since
$$r^2={\bf x^Tx}\quad\Rightarrow\quad\frac{d}{dt}r^2=\frac{d}{dt}{\bf x^Tx}\quad\Rightarrow\quad r\dot{r}=-{\bf u^Tx}$$
measurements of both $r$ and $\dot{r}$ (e.g. from a Doppler sensor)  can be used in the LTV Kalman filter framework, in which case
$$y=r\dot{r}={\bf Hx} $$$$ \bf H=u^T$$
in both 2D and 3D.\\\\

\noindent\textbf{Remark I}

The estimated landmark positions are based on the azimuth model in the
inertial coordinate system fixed to the robot. Thus, the positions of the
landmarks are positions relative to the robot rather than global
locations. Denoting the states of the visible landmarks by ${\bf
  x}_{il}$ , and corresponding measurements $\theta_i$ and $r_i$, each
with independent covariance matrix $P_{i}$.
$$
\dot{\hat{{\bf x}}}_{il}=-{\bf u}-{\Omega}{\bf{\hat{x}}}_{il}+{\bf K(y-H}_{il}{\bf \hat{x}}_{il})$$
$$
\dot{{\bf P}}_i={\bf Q-P}_i{\bf H}_{il}^T{\bf R}^{-1}{\bf H}_{il}{\bf P}_i-\Omega{\bf P}_i+{\bf P}_i\Omega$$

Since the relative landmark positions are conditioned on the local inertial coordinates of the robot, covariances on each pairs of landmarks are fully decoupled, which shrinks the covariance matrix down to the dimension of single landmark’s coordinates. And complexity in all local cases scales linearly with the number of landmarks.

\noindent\textbf{Remark II}

Our cases proposed above provide suggestions to exploit information from different sensor measurements from various instruments under the same framework of LTV Kalman filter. For example, a Doppler radar or sonar measures quite good $r$ and $\dot{r}$; a camera measures the bearings $\phi$ and $\psi$, while combined with optical flows and other algorithms, it can also measure $\dot{\phi}$, $\dot{\psi}$ and $\tau$; and a lidar measures good $r$ and $\phi$ and $\psi$.

\subsection{LTV Kalman filter in 2D global coordinates}

Before introducing the LTV Kalman filter in 2D global coordinates, we first analyze the Kalman filter in an intermediate local coordinate system $C_L$.
\subsubsection{LTV Kalman filter in 2D rotation only coordinates}
The local coordinate $C_L$ is fixed to the origin of the robot at $t=0$, and has the same attitude as the robot, which means $C_L$ has no translation movement but rotates as same as the robot. Here we use the bearing only case (Case I) for example to explain the structure, it can be easily extended to other measurement models. Recall:

$$
\theta=arctan \big(\frac{x_1}{x_2}\big) $$$$
{\bf h}=(\cos\theta, -\sin\theta)$$$$
\Omega=\begin{bmatrix}
0&-\omega\\\omega&0
\end{bmatrix}$$
We name the local coordinates in $C_L$ for each landmark's position ${\bf x}_{iL}$ and the vehicle's position ${\bf x}_{vL}$. For the bearing only case we have the linear constraint
$${\bf h(x}_{iL}-{\bf x}_{vL})=0 
$$
For the kinematics of the system we have both linear velocity and angular velocity on the vehicle and angular velocity alone on the landmarks:
$$
{\bf\dot{x}}_{vL}={\bf u}+\Omega{\bf x}_{vL}
$$
$$
{\bf\dot{x}}_{iL}=\Omega {\bf x}_{iL}
$$

So similar to the LTV Kalman filter proposed in $C_l$ before, we can use another LTV Kalman system updated as:
$${\bf\begin{bmatrix}
{\bf \dot{\hat{x}}}_{1L}\\
{\bf \dot{\hat{x}}}_{2L}\\\vdots\\{\bf \dot{\hat{x}}}_{nL}\\{\bf \dot{\hat{x}}}_{vL}
\end{bmatrix} }= \begin{bmatrix}
 0\\
0\\\vdots\\0\\\bf u
\end{bmatrix}+diag(\Omega){\bf\begin{bmatrix}
{\bf \hat{x}}_{1L}\\
{\bf \hat{x}}_{2L}\\\vdots\\{\bf \hat{x}}_{nL}\\{\bf \hat{x}}_{vL}
\end{bmatrix} }-{\bf PH}^T {\bf R}^{-1}{\bf H}\begin{bmatrix}
{\bf \hat{x}}_{1L}\\
{\bf \hat{x}}_{2L}\\\vdots\\ {\bf \hat{x}}_{nL}\\{\bf \hat{x}}_{vL}
\end{bmatrix} $$
Here $\bf H= \begin{bmatrix}
 {\bf h}_1&0&\cdots&0&-{\bf h}_1\\
0&{\bf h}_2&\cdots&0&-{\bf h}_2\\\vdots&\vdots&\vdots&\vdots&\vdots\\0&0&\cdots&{\bf h}_n&-{\bf h}_n\
\end{bmatrix}$, and the covariance matrix updates as:
$$
\dot{{\bf P}}={\bf Q-PH}^T{\bf R}^{-1}{\bf HP}+diag(\Omega){\bf P-P}diag(\Omega)$$
This system is contracting while free of attitude of the vehicle (heading angle $\beta$ in 2D case). And since the true positions of landmarks and vehicle are particular solutions to this contracting system, all estimated trajectories are guaranteed to converge to the true trajectory exponentially.

\subsubsection{LTV Kalman filter in 2D global coordinates}
Now we look at the problem of LTV Kalman filter in 2D global coordinates. We call the coordinate system $C_G$, which is fixed at the starting point of the robot. Positions of landmarks in global coordinates are ${\bf x}_{iG}$ and position of the vehicle is ${\bf x}_{vG}$. So a transformation matrix from coordinate system $C_G$ to $C_l$ is simply a rotation matrix ${\bf T}(\beta)$. It means ${\bf x}_{iL}={\bf T}(\beta){\bf x}_{iG}$ and ${\bf x}_{vL}={\bf T}(\beta){\bf x}_{vG}$. Substituting these transformations into the LTV Kalman filter proposed in 4.2.1, we can have a LTV Kalman filter in global coordinates describing the same model using virtual measurements as:
$$
{\bf\begin{bmatrix}
{\bf \dot{\hat{x}}}_{1G}\\
{\bf \dot{\hat{x}}}_{2G}\\\vdots\\{\bf \dot{\hat{x}}}_{nG}\\{\bf \dot{\hat{x}}}_{vG}
\end{bmatrix} }= \begin{bmatrix}
 0\\
0\\\vdots\\0\\u\cos\hat{\beta} \\u \sin\hat{\beta}
\end{bmatrix}-{\bf PH}_G^T {\bf R}^{-1}{\bf H}_G \begin{bmatrix}
{\bf \hat{x}}_{1L}\\
{\bf \hat{x}}_{2L}\\\vdots\\{\bf \hat{x}}_{nL}\\{\bf \hat{x}}_{vL}
\end{bmatrix} 
$$ 
Where $${\bf H}_G=\bf \begin{bmatrix}
{\bf h}_1{\bf T}(\hat{\beta})&0&\cdots&0&-{\bf h}_1{\bf T}(\hat{\beta})\\
0&{\bf h}_2{\bf T}(\hat{\beta})&\cdots&0&-{\bf h}_2{\bf T}(\hat{\beta})\\\vdots&\vdots&\vdots&\vdots&\vdots\\0&0&\cdots&{\bf h}_n{\bf T}(\hat{\beta})&-{\bf h}_n{\bf T}(\hat{\beta})\
\end{bmatrix}$$
In this LTV Kalman filter, we extract the attitude of the robot from of the state vector and treat it as a component in the virtual measurement, which helps eliminate the nonlinearity in the model. The value of $\beta$ is determined based on the model by tracking a desired value $\beta_d$,
$$\dot{\hat{\beta}}=\omega+(\beta_d-\hat{\beta})$$
where $\beta_d$ minimizes the quadratic residue error,
$$\beta_d=argmin_{\beta_d\in[-\pi,\pi]}{\bf x}^T{\bf H}^T{\bf Hx}$$
In the 2D case, 
\begin{equation}
\resizebox{\hsize}{!}{$\beta_d=\frac{1}{2}\arctan\frac{\sum_i[-\sin 2\theta_i(x_{iG1}^2-x_{iG2}^2)+2\cos 2\theta_ix_{iG1}x_{iG2}] }{\sum_{i}[-\cos 2\theta_i(x_{iG1}^2-x_{iG2}^2)+2x_{iG1}x_{iG2}\sin 2\theta_i]}\nonumber$}
\end{equation}
when at every instant the sums are taken over the visible landmarks. Since this LTV Kalman filter is obtained simply by a coordinate transformation from the Kalman filter proposed in 4.2.1, which is contracting regardless of the attitude states, this LTV Kalman filter in the global coordinate system is contracting as well exponentially towards the true states.

Because all cases analyzed in previous sections using different
combinations of sensor information follow the same framework, they can
all be treated in global coordinates as same as the bearing only
case. In each of these cases ${\bf H}_G={\bf H}_L diag({\bf T}(\hat{\beta}))$, with a 
different $\beta_d$ minimizing the residue error
${\bf (y-Hx)}^T{\bf (y-Hx)}$ in each case.

Note that for estimating the positions of landmarks and the vehicle in
global coordinates, we are actually utilizing a full state Kalman
filter, so that, computationally, the proposed LTV Kalman filter takes
as much computation as traditional EKF methods. However, our LTV
Kalman filter is linear, global and exact.  Furthermore, it uses a
common framework to solve problems involving different combinations of
sensor information, and contracts exponentially to the true states.
\\\\
\noindent {\bf Remark I: Nonlinearity in vehicle kinematics}

When traditional EKF SLAM methods are applied to ground vehicles, another nonlinearity arises from the vehicle kinematics. This is easily incorporated in our model.

The vehicle motion can be modeled as 
$$\dot{x}_{v1}=u \ \cos\beta $$$$\dot{x}_{v2}=u \ \sin\beta\nonumber$$$$\dot{\beta}=\ \omega\ =\frac{u}{L} \ \tan\theta_s
$$
where $u$ is the linear velocity, $L$ is the distance between the front and rear axles and $\theta_s$ is the steering angle. As the heading angle $\beta$ is an independent input generated by
an upstream level of the filter dynamics, the kinematics of the vehicle remains linear time-varying.
$$
\frac{d}{dt}\begin{bmatrix}
x_{v1}\\
x_{v2}
\end{bmatrix} = 
\begin{bmatrix}
u \ \cos\beta \\ u \ \sin\beta \\
\end{bmatrix}
$$
\\\\

\noindent {\bf Remark II: 3D capability}

It is obvious that our proposals in all cases in local coordinates have full capability of dealing with 6DOF problem. That means when based on the local coordinate system fixed on the robot, or based on a local coordinate system fixed at the origin rotating with the robot, we can deal with 3D problems very easily. Such scenarios are more popular for the flying vehicles to map the surrounding environment rather than large scale long history global mapping. For large scale global mapping, LTV Kalman filter proposed in 4.2.2 has been fully proved to work in 2D applications. In 3D(6DOF) scenarios, the LTV Kalman filter by itself still works with no question, it is just that we may need some other nonlinear optimization methods to find the desired attitude states for the estimated states to track to minimize the residue error. And further we put the estimated value of the attitudes into the LTV Kalman filter as measurements as same as the 2D case.  

\noindent {\bf Remark III: Second-order vehicle dynamics}

The algorithm can be easily extended if instead of having direct velocity measurement, the vehicle dynamics model is second-order,
$${\bf\ddot{x}}_{vG}={\bf u}$$
where $\bf u$ is now the translational \emph{acceleration} instead of the translational velocity, and the state vector is augmented by the linear velocity vector of the vehicle. 
Cases I to IV extend straightforwardly since the constraints
$${\bf h}_i{\bf T}({\bf x}_{iG}-{\bf x}_{vG})=0$$$${\bf h}_i^*{\bf T}({\bf x}_{iG}-{\bf x}_{vG})=r_i\ $$$$(\dot{\theta}_i{\bf h}_i^*+{\bf h}_i\Omega){\bf T}({\bf x}_{iG}-{\bf x}_{vG}) ={\bf -h}_i{\bf\dot{x}}_{vG}\ $$$$| \tau {\bf h}_i^*{\bf\dot{x}}_{vG}| = {\bf h}_i^*{\bf T}({\bf x}_{iG}-{\bf x}_{vG})$$
\noindent remain linear.
Only case V (range only) does not, as it relies on the product of the position and velocity of the vehicle in the model.

\subsection{Contraction analysis for the LTV Kalman filter}
Since all our cases follow the same LTV Kalman filter structure, we can analyze the contraction property in general for all cases at the same time. The LTV Kalman filter system we proposed previously contracts according to Section 3, with metric ${\bf M}_i = {\bf P}_i^{-1}$, as analyzed in \cite{lohmiller2013contraction}:
$$
\frac{\partial {\bf f}_i^T}{\partial {\bf x}_i}{\bf M}_i+{\bf M}_i\frac{\partial {\bf f}_i}{\partial {\bf x}_i}+\dot{{\bf M}}_i
=-{\bf M}_i{\bf Q}_i{\bf M}_i-{\bf H}_{il}^T{\bf R}^{-1}{\bf H}_{il}\nonumber\
$$
The above leads to the global exponential Kalman observer of landmarks (lighthouses) around a vehicle. Hence for any initial value, our estimation will converge to the trajectory of the true landmarks positions exponentially. It gives stability proof to the proposed LTV Kalman filter and boundedness of $\bf M$ is given with the observability grammian. However, LTV Kalman cannot compute the convergence rates explicitly, because the convergence rate is given by the eigenvalues of ${\bf -M}_i{\bf Q}_i{\bf M}_i - {\bf H}_{il}^T{\bf R}^{-1}{\bf H}_{il}$ which is related to $\bf M$.

This system is contracting with metric ${\bf M=P}^{-1}$ in global coordinates also. Since the true locations of landmarks and path of the vehicle are particular solutions to the system, all trajectories of the state vectors would converge exponentially to the truth. 
%
%
%

\subsection{Noise analysis}
A basic assumption for the Kalman filter is that the noise signal
${\bf v}(t)= \bf y-Hx$ is zero-mean. Since the actual measurements
obtained from a robot are $\theta$, $\phi$, $\dot{\theta}$, $\dot{\phi}$, $r$, and $\tau$, we need to verify
that the mean of noise remains zero after incorporating them into the
virtual measurements to transform direct errors of measurements to Cartesian errors between estimation and truth. Similarly, the variance estimates in the initial
noise model have to be ported to the new variables. The general philosophy of this
paper is that typically the noise models themselves are somewhat coarse estimates,
so that this translation of estimated noise variances to the new variables can be approximate
without much practical loss of performance. Recall also that the LTV Kalman filter is the
optimal least-squares LTV filter given the means and variances of the driving and measurement
noise processes, regardless of the noise distribution. In addition, the precision on $\bf Q$ and $\bf R$ does not affect the filter's stability and convergence rates, but only its optimality.

For our proposed algorithm using LTV Kalman filter on SLAM problem, the measurement model can be generally formulated as 
$${\bf v}(t)= {\bf y}-{\bf H}(\theta,\phi,\dot{\theta},\dot{\phi}){\bf x}$$ 
For components of the measurement $\bf y$, they can be both virtual measurements like 
$$0={\bf h}{\bf x}$$
$$(\dot{\theta}{\bf h^*+h}\Omega){\bf x =-hu}$$
$$|\tau {\bf h^*u}|={\bf h}^*{\bf x}$$
$$r\dot{r}={\bf u}^T{\bf x}$$
and actual measurements like 
$$r={\bf h}^*{\bf x}$$

And noise come from both the measurements $y$ and the model matrix ${\bf H}(\theta,\phi,\dot{\theta},\dot{\phi})$, since the model matrix takes noisy measurements of $\theta,\phi,\dot{\theta},\dot{\phi}$ as inputs. For the measurements part, if it comes from actual measurement, then noise analysis comes directly from sensor specifications. If it comes from virtual measurements, it is easy to use different methods like the Monte Carlo to calibrate mean and variance of the virtual measurement noises. Therefore, in this section we focus on analyzing noises that come from model matrices. More specifically, we discuss about noise from
$${\bf hx}\quad \quad\text{and}\quad\quad{\bf h}^*{\bf x}$$

For Cases I, II, III and IV, assume the bearing angle $\theta$ we measure comes with a zero-mean white Gaussian noise $$w_{\theta}\sim N(0,\sigma_{\theta}^2)$$ and $\phi$ with a zero-mean white Gaussian noise $$w_{\phi}\sim N(0,\sigma_{\phi}^2)$$
then 
\begin{eqnarray}
E[\cos(\theta +w)]&=&E[\cos\theta\cos(w)-\sin\theta\sin(w)]\nonumber\\
&=&\cos\theta E[\cos(w)]-\sin\theta E[\sin(w)]\nonumber\\
&=&e^{-\frac{\sigma_\theta^2}{2}}\cos\theta \nonumber
\end{eqnarray}
and similarly 
\begin{eqnarray}
E[\sin(\theta +w)]&=&E[\cos\theta\sin(w)+\sin\theta\cos(w)]\nonumber\\
&=&\cos\theta E[\sin(w)]+\sin\theta E[\cos(w)]\nonumber\\
&=&e^{-\frac{\sigma_\theta^2}{2}}\sin\theta \nonumber
\end{eqnarray}

Combined with the geometry:

\noindent (2D)
$$x_1=r \sin\theta $$$$	x_2= r \cos\theta$$
\noindent (3D)
$$x_1=r \sin\theta\sin\phi $$$$ 	x_2= r \cos\theta\sin\phi $$$$ x_3= r \sin\phi$$
For our virtual measurement $$y={\bf hx}+v$$ 

\noindent where $${\bf h}=[\cos\theta, -\sin\theta]$$ the noise $$v=0-{\bf hx}=-(x_1 \cos\theta -x_2 \sin\theta)$$ 
so that the mean of the noise 
\begin{eqnarray}
E[v]&=&E[0-{\bf hx}]\nonumber\\
&=&-e^{-\frac{\sigma_\theta^2}{2}}(x_1 \cos\theta-x_2 \sin\theta)\nonumber\\
&=&0\nonumber\\
\end{eqnarray}
which means there is no bias in this case.

In the 3D case 

$$
{\bf h} = \left( \begin{array}{ccc} \cos \theta & -sin \theta & 0 \\ 
-\sin \phi \sin \theta & -\sin \phi \cos \theta & \cos \phi \end{array} \right)
$$
\begin{eqnarray}
E[\bf{v}]&=&E[0-{\bf hx}]\nonumber\\
&=&\left(\begin{array}{c}-e^{-\frac{\sigma_\theta^2}{2}}(x_1 \cos\theta -x_2 \sin\theta)\\
r(e^{-\frac{\sigma_\phi^2}{2}}-e^{-\frac{\sigma_\theta^2+\sigma_\phi^2}{2}})\cos\phi \sin\phi
 \end{array}\right)\nonumber\\
 &=&\left(\begin{array}{c}0\\
-r(e^{-\frac{\sigma_\phi^2}{2}}-e^{-\frac{\sigma_\theta^2+\sigma_\phi^2}{2}})\cos\phi \sin\phi
 \end{array}\right)\nonumber
\end{eqnarray}

So there would be a small bias in the second term.

And for virtual measurement $$y={\bf h^*x}+v$$ 

\noindent where $${\bf h^*}=[\sin\theta, \cos\theta]$$ the noise $$v=r-{\bf h^*x}=r-(x_1 \sin\theta -x_2 \cos\theta)$$

\noindent so that the mean of the noise 
\begin{eqnarray}
E[v]&=&E[r-{\bf h^*x}]\nonumber\\
&=&r-e^{-\frac{\sigma_\theta^2}{2}}(x_1 \sin\theta+x_2 \cos\theta)\nonumber\\
&=&(1-e^{-\frac{\sigma_\theta^2}{2}})r\nonumber\\
\end{eqnarray}

which means there is small bias in mean value, where in the 3D case
\begin{eqnarray}
E[v]&=&E[r-{\bf h^*x}]\nonumber\\
&=&r-e^{-\frac{\sigma_\theta^2}{2}}(e^{-\frac{\sigma_\phi^2}{2}}x_1 \sin\theta \sin\phi+e^{-\frac{\sigma_\phi^2}{2}} x_2 \cos\theta \sin\phi+x_3 \cos\phi)\nonumber\\
&=&r(1-e^{-\frac{\sigma_\phi^2}{2}}\sin^2\phi-e^{-\frac{\sigma_\theta^2+\sigma_\phi^2}{2}}\cos^2\phi)\nonumber\\
\end{eqnarray}

\begin{figure*}[htbp] 
\centering 
\centerline{\includegraphics[width=1.2\textwidth]{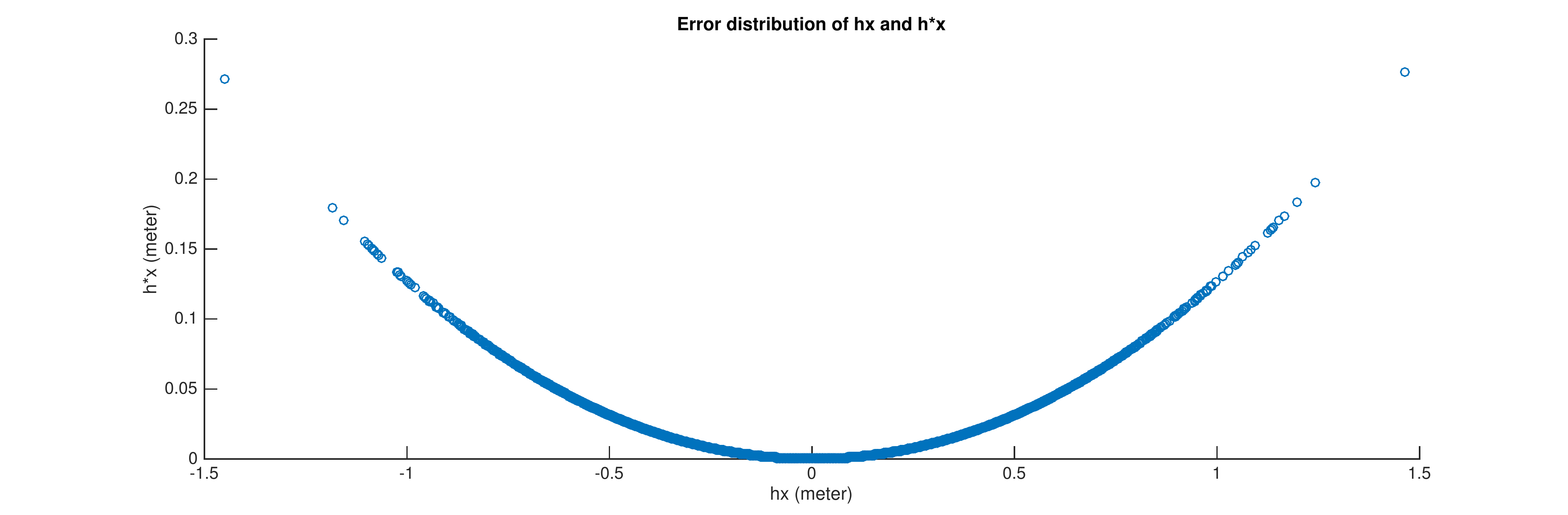}}

\caption{Error distribution of ${\bf hx}$ and ${\bf h}^*{\bf x}$}
\label{errorbias}
\end{figure*}

\begin{figure*}[htbp] 
\centering 
\centerline{\includegraphics[width=1.2\textwidth]{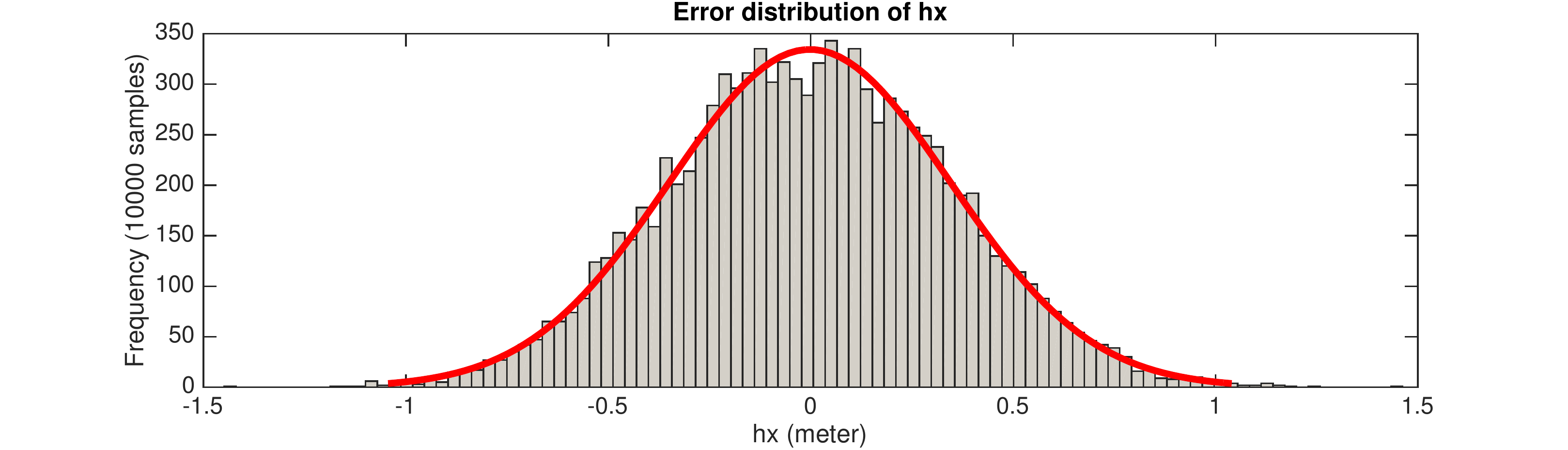}}

\caption{Error distribution of ${\bf hx}$}
\label{errorhx}
\end{figure*}

Besides theoretical analysis about the noises of ${\bf hx}$ and ${\bf h}^*{\bf x}$, we also provide simulation results supporting the analysis.
We set up a simple 2D simulation environment as $r=4$m, $\sigma_r=0.2$m, $\theta=45^\circ$ and $\sigma_{\theta}=5^\circ$. We use 10000 samples to estimate and analyze the errors. Fig. \ref{errorbias} shows the plot of ${\bf hx}$ and ${\bf h}^*{\bf x}$ combined. More specifically, we can see from the histogram of ${\bf hx}$ in Fig. \ref{errorhx} that the ported noise is bias free, and close to a Gaussian distribution. Combined with the Monte Carlo method we get that the mean-shift of 10000 samples is $-0.0016$, which supports the analysis that the noise stays bias free. From histogram of ${\bf h}^*{\bf x}$, error distribution of ($r_{real}-{\bf h}^*{\bf x}$) is one-sided, which means there is a bias that makes the noise not zero-mean. We then plot the change of noise distribution between the original range measurement $\sigma_r$ and the ported noise $r_{measured}-{\bf h}^*{\bf x}$ as shown in Fig. \ref{meanshift}. The means-shift is very small under the setting and negligible. The Monte Carlo result about the mean shift of 10000 samples is $0.0151$, which is consistent with our analytical result of the bias $(1-e^{-\frac{\sigma_\theta^2}{2}})r=0.0152$ when we substitute in the numbers. Such a bias is small enough to ignore considering that $\sigma_r=0.2$m. 

\begin{figure*}[htbp] 
\centering 
\centerline{\includegraphics[width=1.2\textwidth]{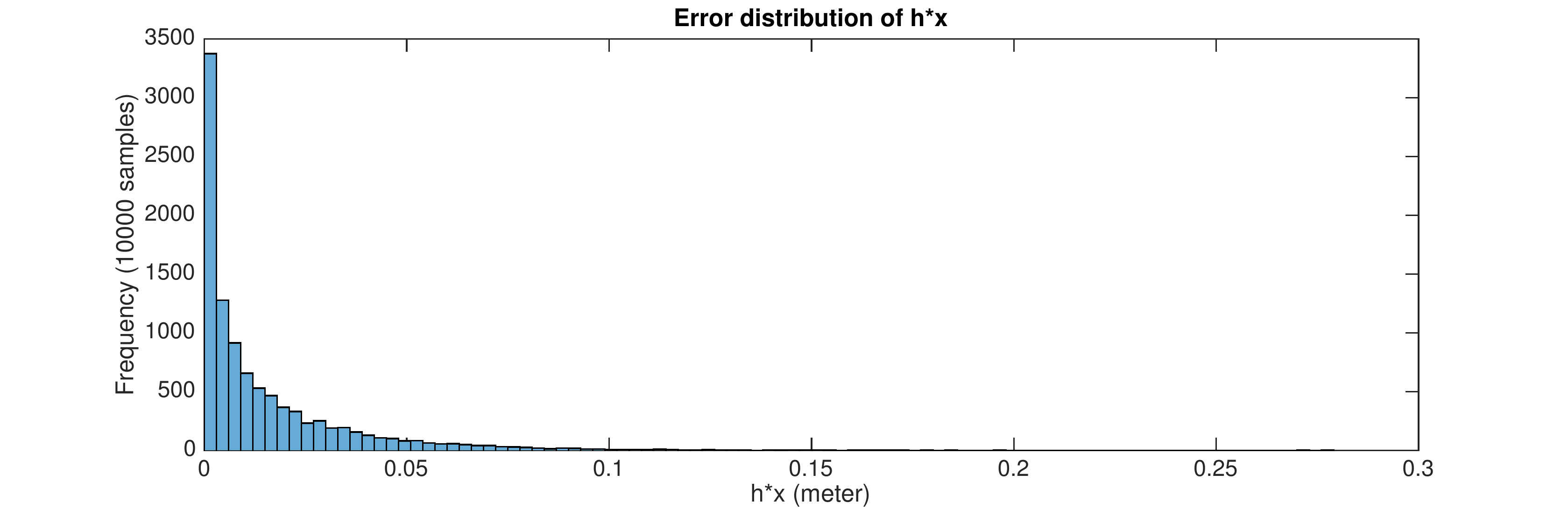}}

\caption{Error distribution of $r_{real}-{\bf h}^*{\bf x}$}
\label{errorhsx}
\end{figure*}

\begin{figure*}[htbp] 
\centering 
\centerline{\includegraphics[width=1.2\textwidth]{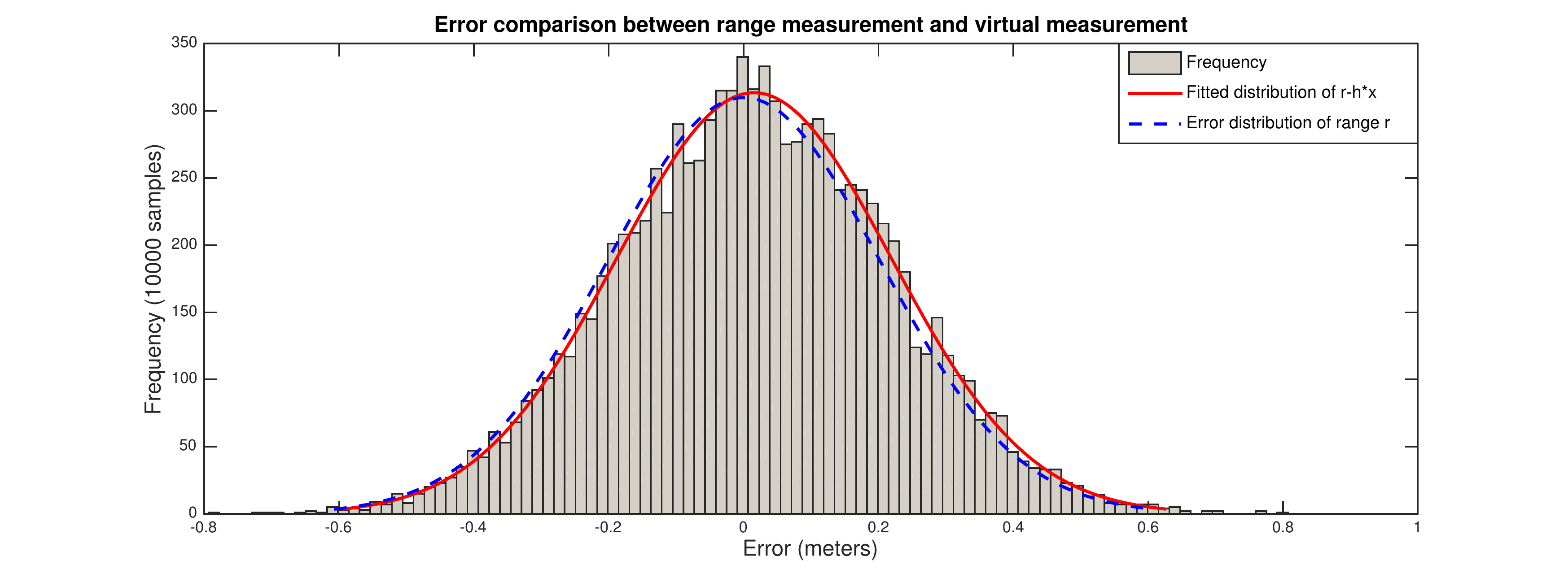}}

\caption{Error comparison about range measurement between actual and virtual measurements $r$ and $r={\bf h}^*{\bf x}$}
\label{meanshift}
\end{figure*}

Following the same logic and process, we get that in Case II (3D)
$$
E[\bf{v}]=\left(\begin{array}{c}0\\
-r(e^{-\frac{\sigma_\phi^2}{2}}-e^{-\frac{\sigma_\theta^2+\sigma_\phi^2}{2}})\cos\phi \sin\phi\\
r(1-e^{-\frac{\sigma_\phi^2}{2}}\sin^2\phi-e^{-\frac{\sigma_\theta^2+\sigma_\phi^2}{2}}\cos^2\phi)
 \end{array}\right)
$$

In Case III (3D), 
\begin{equation}
E[\bf{v}]=\left(\begin{array}{c}0\\
-r(e^{-\frac{\sigma_\phi^2}{2}}-e^{-\frac{\sigma_\theta^2+\sigma_\phi^2}{2}})\cos\phi \sin\phi\\
\dot{\theta}(e^{-\frac{\sigma_\theta^2}{2}}-e^{-\frac{\sigma_\phi^2}{2}})r\sin^2\phi+(e^{-\frac{\sigma_\theta^2}{2}}-e^{-\frac{\sigma_\theta^2+\sigma_\phi^2}{2}})r\cos^2\phi)\\
(e^{-\frac{\sigma_\phi^2}{2}}-e^{-\frac{\sigma_\theta^2+\sigma_\phi^2}{2}})(u_3\cos\phi-\dot{\phi}r\sin^2\phi)
 \end{array}\right)\nonumber
\end{equation}
In Case IV (3D), 
$$
E[\bf{v}]=\left(\begin{array}{c}0\\
-r(e^{-\frac{\sigma_\phi^2}{2}}-e^{-\frac{\sigma_\theta^2+\sigma_\phi^2}{2}})\cos\phi \sin\phi\\
(e^{-\frac{\sigma_\phi^2}{2}}-e^{-\frac{\sigma_\theta^2+\sigma_\phi^2}{2}})(\tau u_3 \sin\phi-r\sin^2\phi)
 \end{array}\right)
$$

We can see that in each case, the means of noise only shifts with a scale coefficient of $(e^{-\frac{\sigma_\phi^2}{2}}-e^{-\frac{\sigma_\theta^2+\sigma_\phi^2}{2}})$ or $(1-e^{-\frac{\sigma_\theta^2}{2}})$. When the variances $\sigma_\theta$ and $\sigma_\phi$ are small, that coefficient is almost zero. Even when we increase in simulations the actual variances of the bearing measurements to $10^{\circ}$ (which is unrealistic based on the performances of current instruments), the mean shift is still in on the scale of $10^{-2}$m. It thus remains negligible and does not need to be subtracted.

For Case V, since $\bf H=u^T$ and radius $r$ are measured independently, there would be no mean-shift for the noise $$v={\bf u^Tx}+r\dot{r}$$ which means $$E[v]=0$$ in both 2D and 3D cases. 

\noindent {\bf Remark}\\\\

The variance of the noise on the virtual measurements can also be
easily approximated. For the bearing and range virtual measurements, one has
\begin{eqnarray}
{\bf hx}&=&r(\cos(\theta+w)\sin(\theta)-\sin(\theta+w)\cos(\theta))\nonumber\\
&=&-r\sin w\nonumber\\
\end{eqnarray}

\begin{eqnarray}
r-{\bf h}^*{\bf x}&=&r-r(\sin(\theta+w)\sin(\theta)+\cos(\theta+w)\cos(\theta))\nonumber\\
&=&r(1-\cos w)\nonumber\\
&=&2r\sin^2(w/2)\nonumber\\
\end{eqnarray}
In such case
\begin{eqnarray}
Var({\bf hx})&=&E[r^2\sin^2 w]\nonumber\\
&=&r^2E[\sin^2 w]\nonumber\\
&\leq &r^2E[w^2]\nonumber\\
&=& \sigma_{\theta}^2 r^2\nonumber\\
\end{eqnarray}
and 
\begin{eqnarray}
Var(r-{\bf h}^*{\bf x})&=&E[2r\sin^2(w/2)]\nonumber\\
&=&4r^2E[\sin^4 (w/2)]\nonumber\\
&\leq &4r^2E[(w/2)^4]\nonumber\\
&=& \frac{\sigma_{\theta}^4}{4} \ r^2\nonumber\\
\end{eqnarray}

For covariance between the radial and tangential errors, we have
\begin{eqnarray}
Cov(r-{\bf h^*x}, {\bf hx})&=&E[r(1-\cos w)*(-r\sin w)]\nonumber\\
&=&r^2E[\cos w \sin w]-r^2E[\sin w]\nonumber\\
&=&0 \nonumber\\
\end{eqnarray}
Since the exact $r$ is not known, when computing matrix $R$ it may be conservatively
replaced by a known upper bound $r_{max}$ , or more finely
by $$r^*=min(r_{measured}+3\sigma_r\ ,\ r_{max})$$ where $\sigma_r$ is the variance of
the range measurement noise.

\subsection{Extensions}
\begin{figure*}[htbp] 
\centering 
\includegraphics[width=0.6\textwidth]{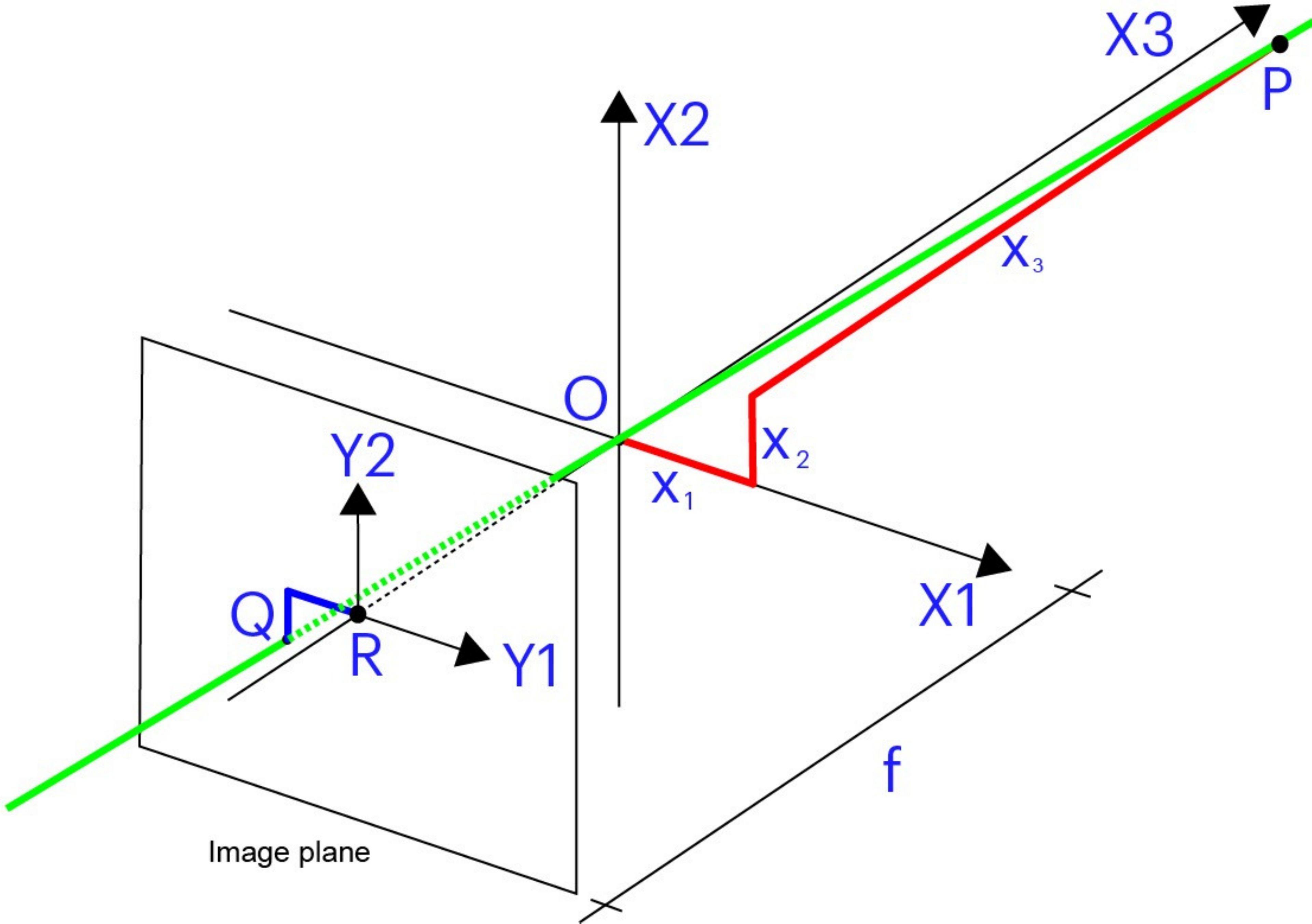}
\caption{Pinhole Camera Model}
\end{figure*}

\noindent{\bf Pinhole camera model}

The principle of transforming a nonlinear measurement into a LTV
representation is applicable to other contexts, for instance the
pinhole camera model (Fig. 2),

$$\begin{bmatrix}
y_1\\y_2
\end{bmatrix}=-\frac{f}{x_3}\begin{bmatrix}
x_1\\x_2
\end{bmatrix}$$

\noindent which can be rewritten as LTV constraints on the states $(x_1, x_2, x_3)$ , 
$$
\begin{bmatrix}
f&0&y_1\\0&f&y_2
\end{bmatrix}\begin{bmatrix}
x_1\\x_2\\x_3
\end{bmatrix}={\bf Hx}={\bf 0}
$$ based on the measured $y_1$ and $y_2$. If in addition we measure
the velocity ${\bf u}$ of the camera center and the angular velocity
matrix $\Omega$ describing the vehicle's rotation, the kinematics
model is
$$\dot{\bf x}=-{\bf u}-\Omega {\bf x}$$
So we can extend the same LTV Kalman system to estimate the local position of the target shown on the image plane as:
$${\bf \dot{\hat{x}}}= -{\bf u}-\Omega{\bf \hat{x}} - {\bf PH}^T{\bf R}^{-1}{\bf H\hat{x}}
$$
with covariance updates
$$
\dot{{\bf P}}={\bf Q-PH}^T{\bf R}^{-1}{\bf HP}-\Omega{\bf P+P}\Omega$$

\noindent{\bf Structure from motion}

Structure from motion \citep{soatto1996motion, soatto1997recursive, soatto1998reducing} is a problem in machine vision field dealing with range imaging. The problem is to recover the 3D model of a structure (building, furniture, etc.) from a series of 2D images. The simplified structure from motion problem can be modeled as a combination of SLAM and pin-hole camera. Intuitively, it may be a global version of the pin-hole camera model we introduced previously, where estimations on local positions of the features are replaced by estimations on global positions of features along with global pose and attitude of the camera. So similar with the discussions in the pin-hole model, we can easily write the LTV constraints about any observation as:
$$
\begin{bmatrix}
f&0&y_{i1}\\0&f&y_{i2}
\end{bmatrix}{\bf T}(\beta)({\bf x}_i-{\bf x}_c)={\bf 0}
$$
where ${\bf T}(\beta)$ is the rotation matrix from global coordinates to local coordinates on the camera, and ${\bf x}_i, {\bf x}_c$ are respectively position of a feature and position of the camera. As in the method we used in the global SLAM discussion, we can have a separate estimation on $\beta$ and treat it as an input to the LTV Kalman filter as:
$$\dot{\hat{\beta}}=\omega+(\beta_d-\hat{\beta})$$
while kinematics of the estimated states are simply:
$$\dot{{\bf x}}_i={\bf 0}$$
$$\dot{{\bf x}}_c={\bf u}$$

\section{Experiments}

\noindent\textbf{Experiments for 2D landmarks estimation}\\
We experiment the 2D version of our cases with simulations in Matlab. As shown in \url{https://vimeo.com/channels/910603}, in the simulations, we have three landmarks. The diameter of each landmark is $d=2$m. We have run simulations on all five cases. The noise signals that we use in the simulations are: standard variance for zero-mean Gaussian noise of $\theta$ is $2\degree$; standard variance for noise of $\dot{\theta}$ is $5\degree$/s; standard variance for measurement noise of $r$ is $2m$; and standard variance for noise of $\alpha$ is $0.5\degree$.
\begin{figure*}[htbp] 
\centering 
\includegraphics[width=0.9\textwidth]{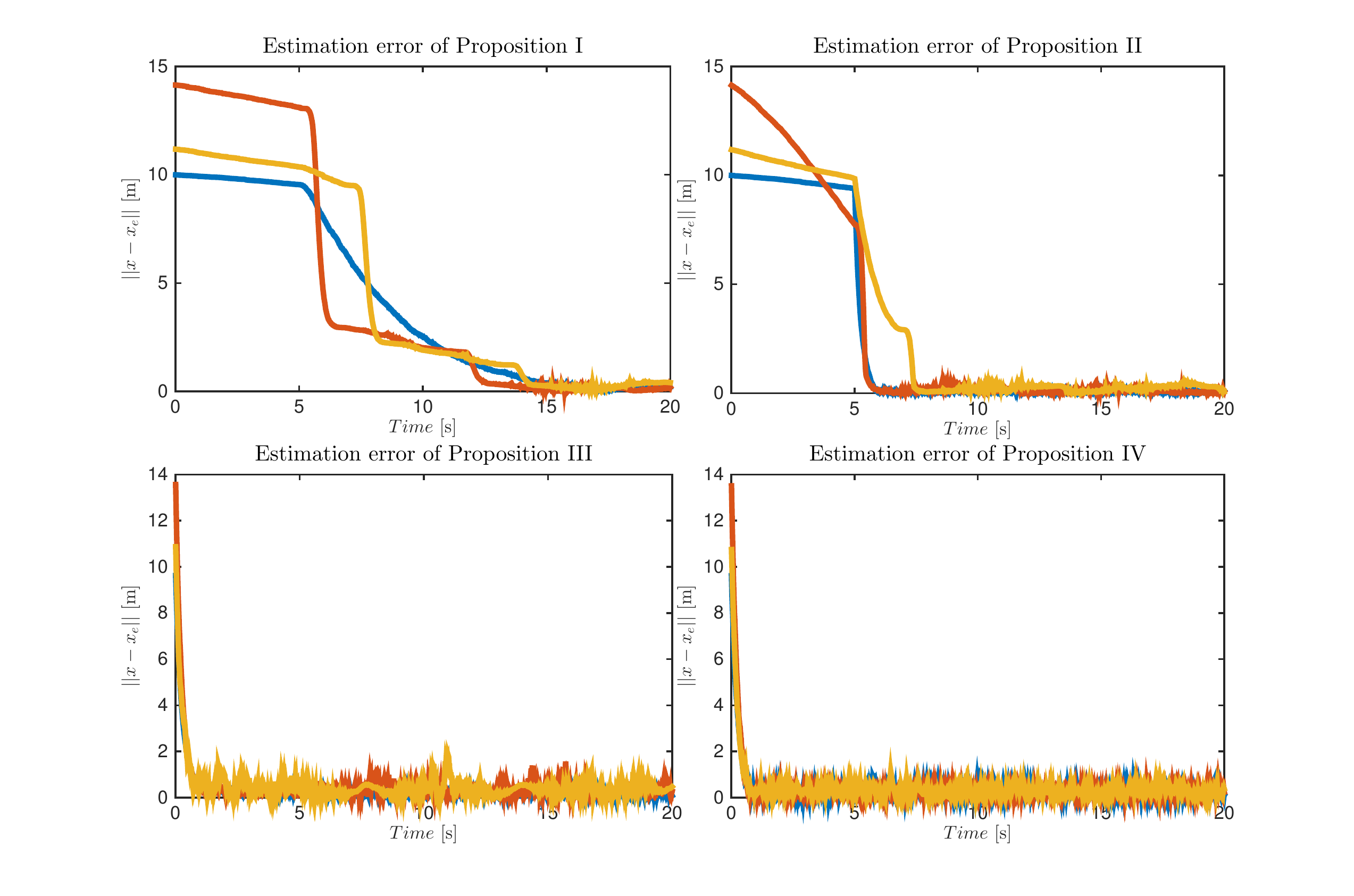}

\caption{Estimation error for Case I, II, III, IV for 2D estimation}
\label{error}
\end{figure*}
\begin{figure*}[htbp] 
\centering 
\includegraphics[width=0.9\textwidth]{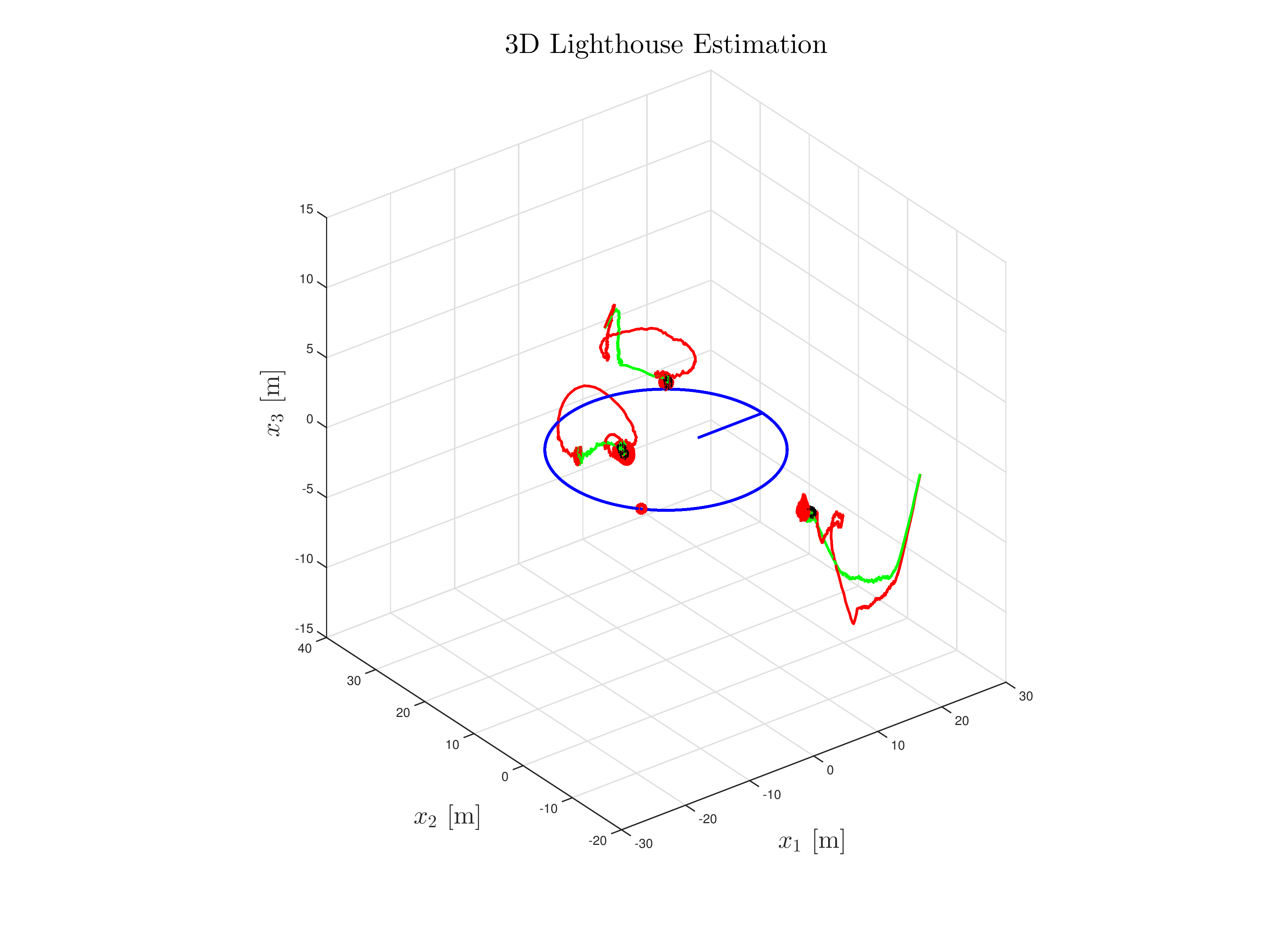}

\caption{3D landmarks estimation in Case I and III}
\label{3d}
\end{figure*}
In \url{https://vimeo.com/channels/910603}, the green lines indicate the trajectories of estimations of the each case. The blue lines are the movement trajectory of the vehicle. As shown in \url{https://vimeo.com/channels/910603}, trajectories of estimations from Case II, III and IV are smoother and converge faster than original Case I and Case V. This is because the trajectory exploits additional information. In particular, for Case II and IV, since the "time-to-contact" measurement and radial distance measurement both contains information on the radial direction, they converge to the true position directly, without waiting for the vehicle movement to bring in extra information.

Next, we analyze the estimation errors $||\bf x-\hat{x}||$ in Case I, II, III and IV for all three landmarks in Fig. \ref{error}. The figure shows that the errors decay faster in Case II, III and IV. The difference is that Case III needs to wait for the movement of the vehicle to provide more information about $\dot{\theta}$, yet Case II and IV contract much faster with exponential rates because of range related measurements. Compared to Case II, Case IV is less smooth, as expected, because the time-to-contact measurement itself is an approximation and may be disturbed when $\dot{r}$ is close to zero. 
\\
\\
\noindent\textbf{Experiments for 3D landmarks estimation}

We also have simulation results for Case I and Case III in 3D settings. Here we also have three lighthouses with different locations. The results shown in Fig. \ref{3d} suggest that our algorithm is capable of estimating landmarks positions accurately in 3D space with bearing angle for both yaw and pitch. Animations of all simulation results are provided at \url{https://vimeo.com/channels/910603}.

\begin{figure*}[htbp] 
\centering 
\includegraphics[width=0.48\textwidth]{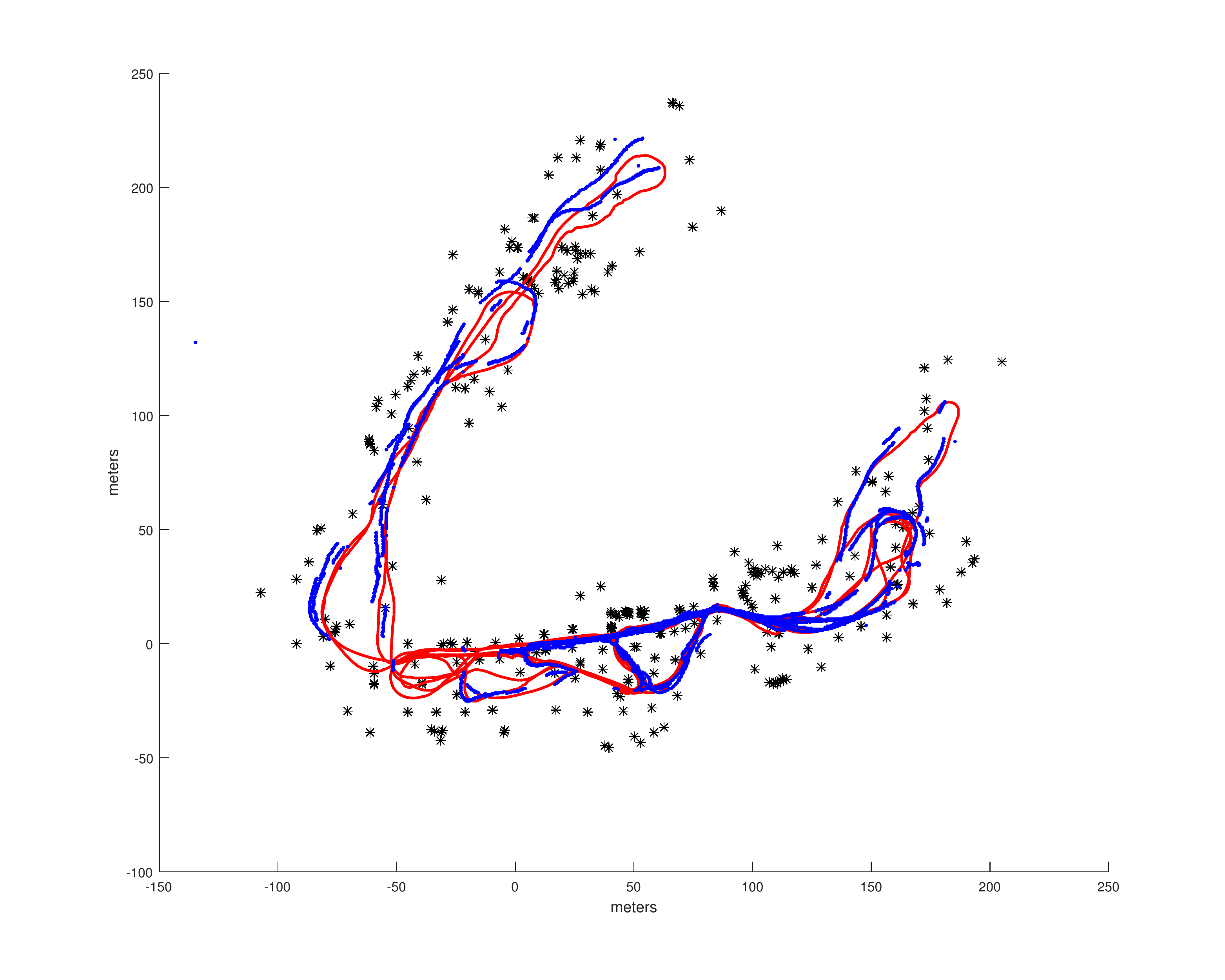}
\includegraphics[width=0.48\textwidth]{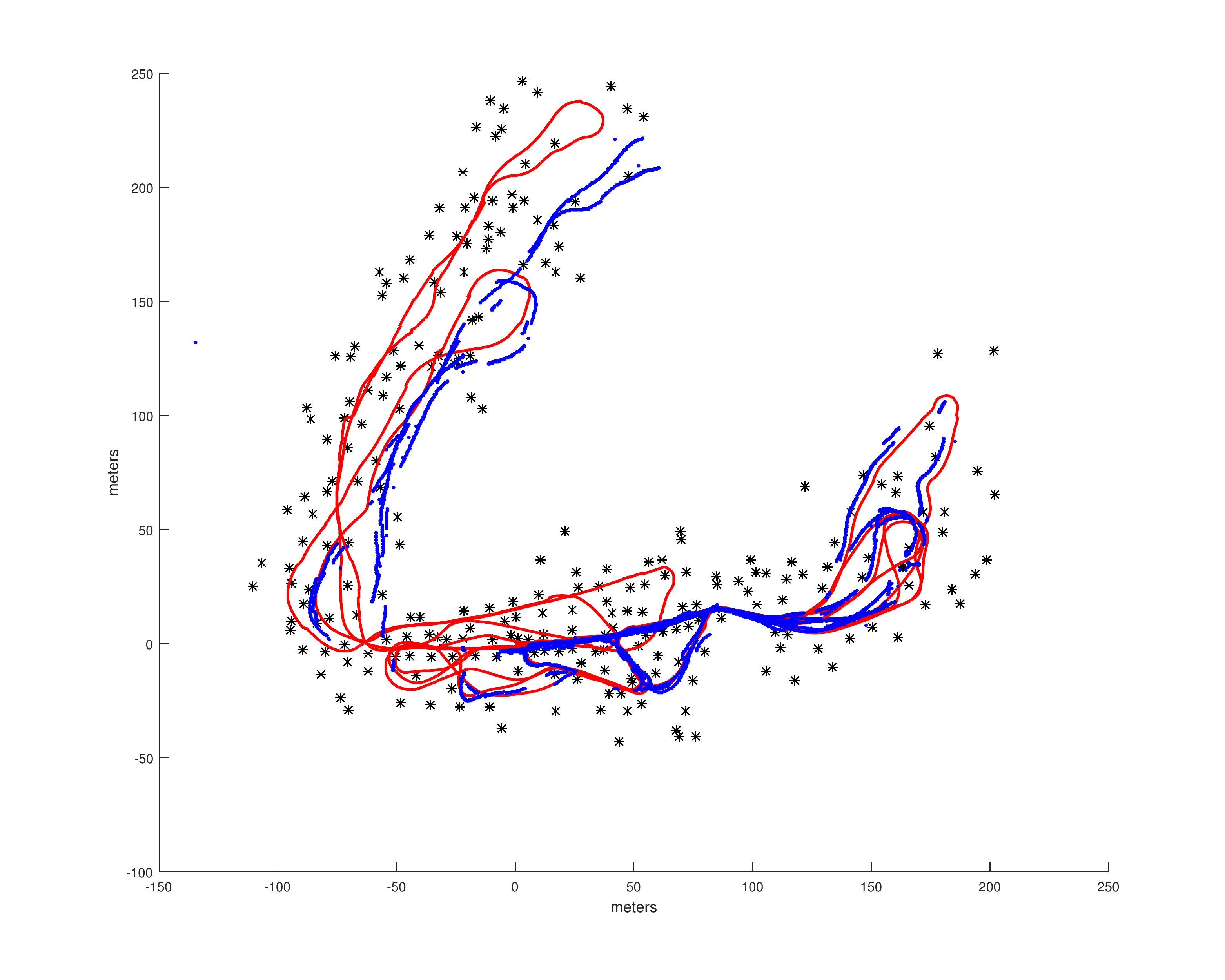}

\caption{ On the left is the path and landmarks estimation of our algorithm and on the right is the result from Unscented Fast SLAM. The thick blue path is the GPS data and the solid red path is the estimated path; the black asterisks are the estimated positions of the landmark.}
\label{vic}
\end{figure*}

%

\noindent\textbf{Experiment for Victoria Park landmarks estimation}
\\
We applied our algorithm to Sydney Victoria Park dataset, a popular dataset in the SLAM
community. The vehicle path around the park is about 30 min, covering over 3.5 km. Landmarks in the park are mostly trees. Estimation results are compared with intermittent GPS information as ground truth to validate the states of the filters as shown in Fig. \ref{vic}. Our estimated track compares favorably to benchmark result of \cite{uslam}, which highlights the consistency of our algorithm in large scale applications. Simulation result of the Victoria Park dataset is provided at \url{https://vimeo.com/136219156}.
\\\\

\section{Decoupled Unlinearized Networked Kalman-filter (SLAM-DUNK)}

\begin{figure*}[htbp] 
\centering 
\includegraphics[width=0.8\textwidth]{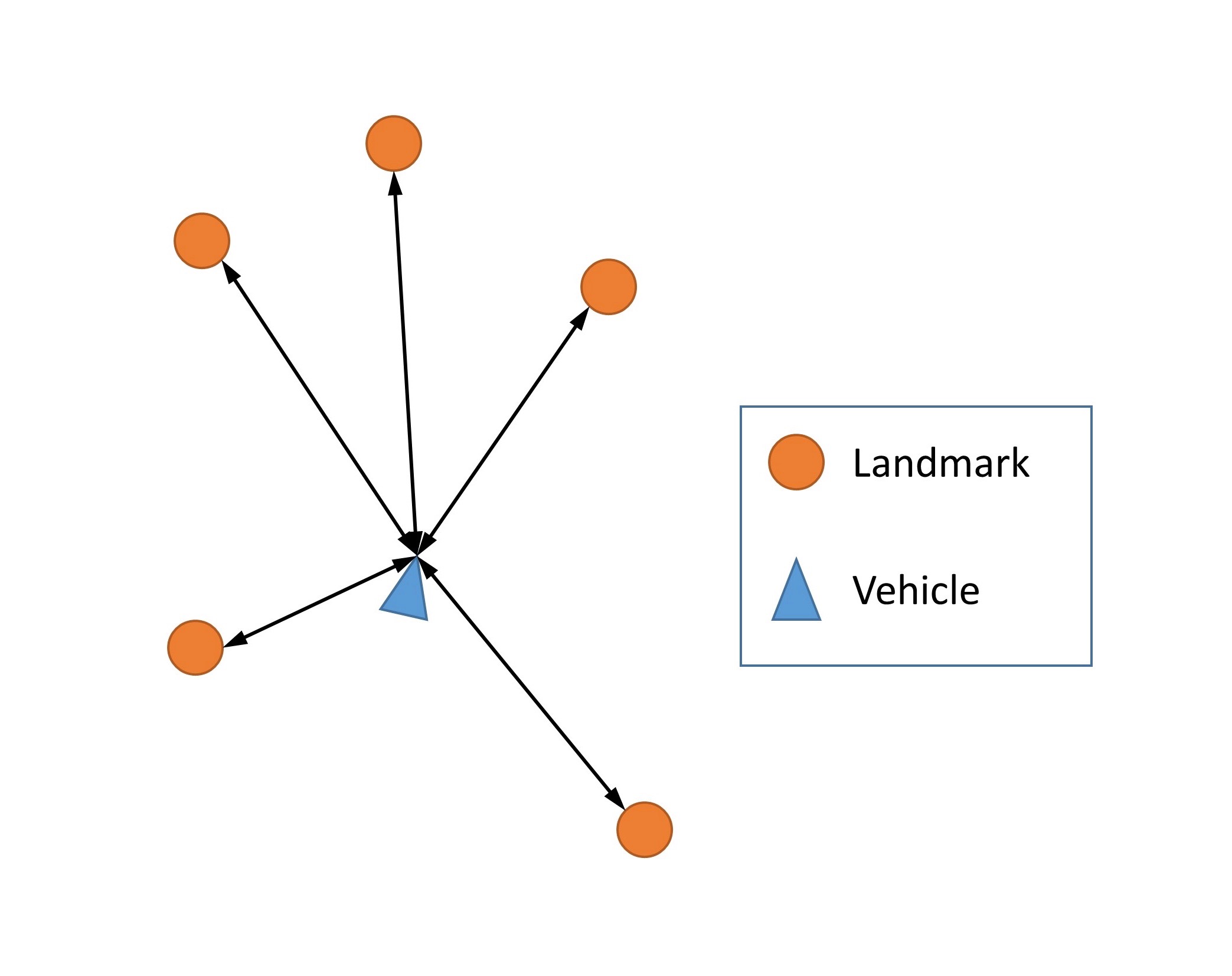}

\caption{ Graph model of EKF-SLAM: all states including both the landmarks and the vehicle are coupled together}
\label{ekf}
\end{figure*}

\begin{figure*}[htbp] 
\centering 
\includegraphics[width=0.8\textwidth]{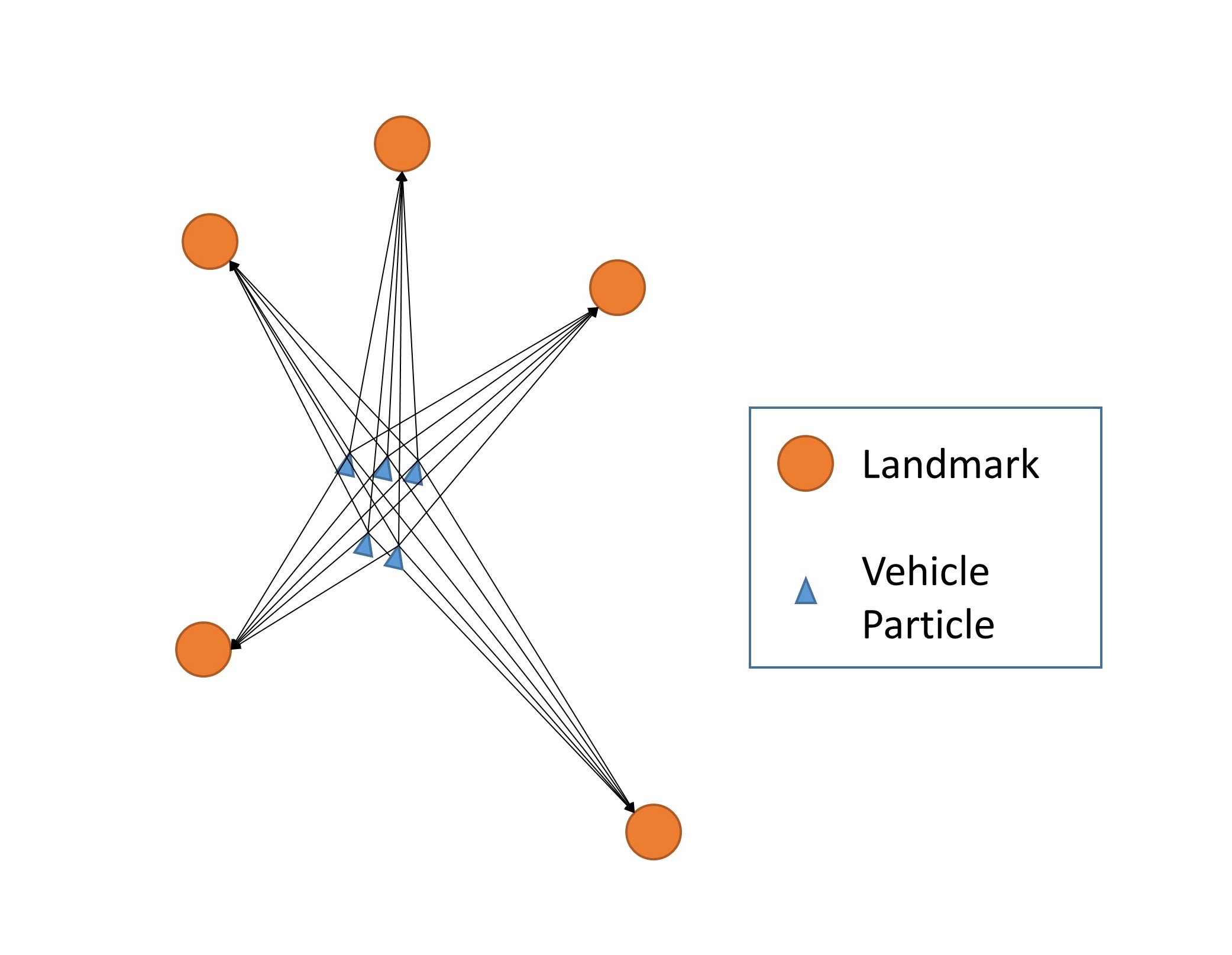}

\caption{ Graph model of Fast-SLAM: states of landmarks are fully decoupled conditioned on each particle of vehicle states}
\label{fast}
\end{figure*}

\begin{figure*}[htbp] 
\centering 
\includegraphics[width=0.8\textwidth]{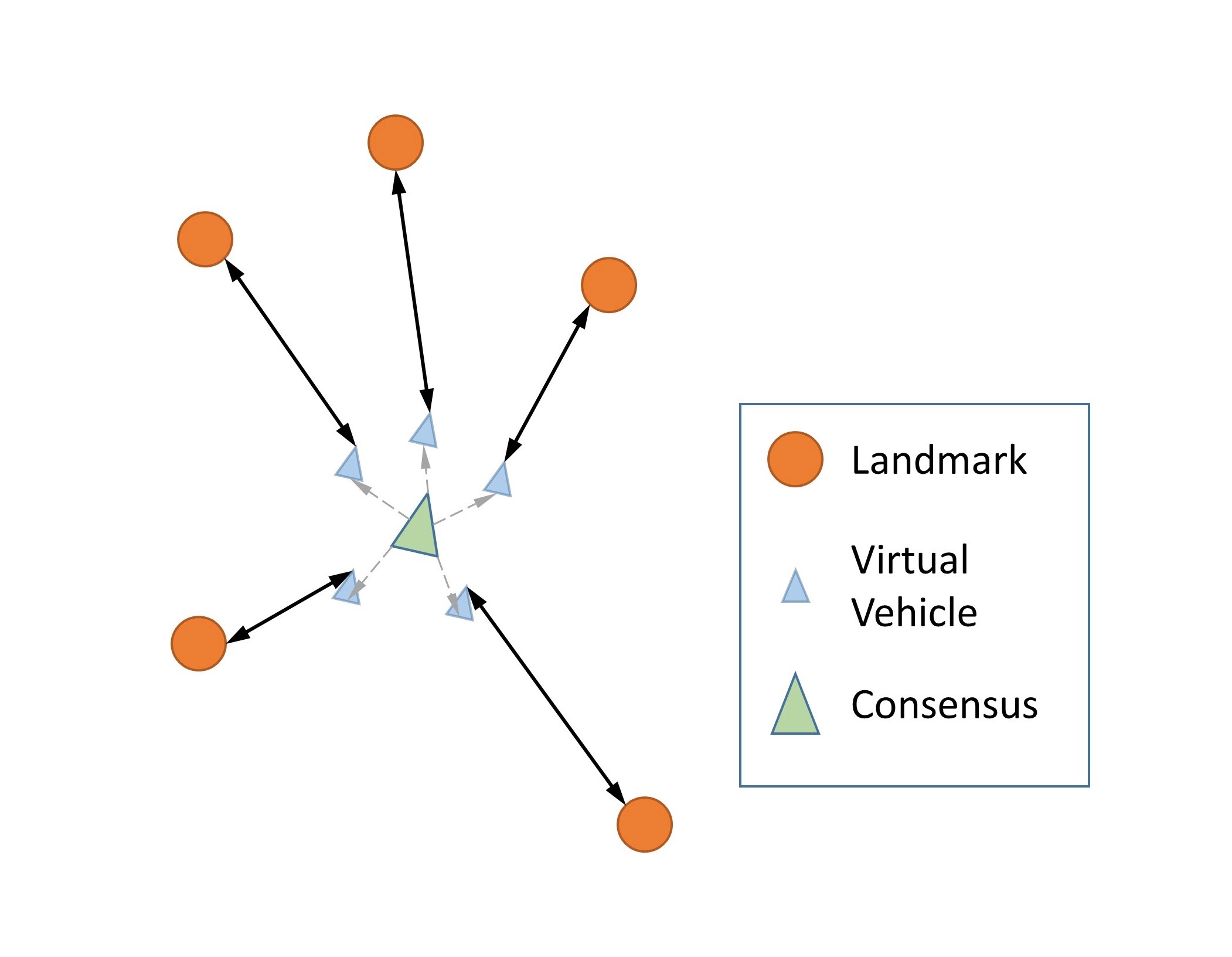}

\caption{ Graph model of SLAM-DUNK: states of each landmark are only coupled with the corresponding virtual vehicle, and consensus of virtual vehicles as maximization of likelihood is used as best estimate}
\label{dunk}
\end{figure*}

One of the main problems for the proposed full LTV Kalman filter in global coordinates is computation complexity. As shown in Fig. \ref{ekf}, since all landmarks are coupled to each other by the vehicle's state, we will have to deal with a full covariance matrix, which requires $O(n^2)$ storage and $O(n^3)$ computation in each step, where $n$ is the number of landmarks. This key limitation restrains the algorithm from being applied to large-scale environment models that could easily contain tens of thousands of features. 
\\
Actually the SLAM problem exhibits important conditional independence: that is, conditioned on the vehicle's states of path, all landmarks are decoupled and independent of each other, as suggested in \cite{montemerlo2002fastslam}. In other words, if we feed the vehicle states estimated by other methods into the filters as prior information, we can decoupled the full LTV Kalman filter into $n$ independent location estimation problems, one for each landmark. For example, in \cite{montemerlo2002fastslam} factorized the SLAM problem into a graph model like Fig. \ref{fast}, where they use $M$ particle filters to update the states of vehicle, and each particle of vehicle is connected to $n$ independent EKF estimators, so that there would be $nM$ filters in total, which means $O(nM)$ computation complexity.
\\
We are proposing a novel algorithm that can decouple the covariance between landmarks into smaller independent estimators and requires less computation even comparing to FastSLAM. Instead of dealing with all measurements and landmarks as one whole state vector to estimate, which is done by full filters like EKF are doing, we want to process information from each single measurement independently with one specific virtual vehicle. Then we establish a consensus summarizing all the information and feedback to the individual observers. Graph model of the proposed algorithm is shown in Fig. \ref{dunk}, and detailed algorithm design is introduced below.
\\

\subsection{Distributed sensing}
For each landmark ${\bf x}_i$, we assign a virtual vehicle ${\bf x}_{vi}$ exclusively to process any information generated from that landmark. Using the case where we have both range and bearing measurements for example, the linear constraints between landmark ${\bf x}_i$ and virtual vehicle ${\bf x}_{vi}$ would be:
$$
{\bf y}_{i1}={\bf H}_{i1} \left[\begin{array}{c}{\bf x}_i\\{\bf x}_{vi} \end{array}\right]
$$
where
$${\bf y}_{i1}=\left[\begin{array}{c} 0\\r_i \end{array}\right]$$ and $${\bf H}_{i1}=\left[\begin{array}{cccc} \sin{\theta_i}&-\cos{\theta_i}&-\sin{\theta_i}&\cos{\theta_i}\\
\cos{\theta_i}&\sin{\theta_i}&-\cos{\theta_i}&-\sin{\theta_i}\end{array}\right]{\bf T}(\beta)$$
\noindent Such constraint is similar with the one we discussed before, with additional rotation term due to global coordinates. In that case, each landmark ${\bf x}_i$ is coupled with a virtual vehicle ${\bf x}_{vi}$ exclusively. Using LTV Kalman filter for each pair, information from observation of any single landmark gets conveyed to the virtual vehicle layer.
$$
\left[\begin{array}{c}\dot{\hat{{\bf x}}}_{i}\\{\dot{\hat{{\bf x}}}_{vi}}\end{array}\right]=\left[\begin{array}{c}0\\{\bf u}\end{array}\right]+{\bf P}_i{\bf H}_i^T{\bf R}^{-1}({\bf y}_i-{\bf H}_i\left[\begin{array}{c}{\hat{{\bf x}}}_{i}\\{{\hat{{\bf x}}}_{vi}}\end{array}\right])$$
$$
\dot{{\bf P}}_i={\bf Q-P}_i{\bf H}_{i}^T{\bf R}^{-1}{\bf H}_{i}{\bf P}_i$$
\\
\subsection{Consensus among virtual vehicles}
In the layer of virtual vehicles, we then summarize information from all observations to get a consensus, and use that consensus to guide all virtual vehicles to follow, which makes the virtual vehicle layer a ``leader-follower'' network. The consensus ${\bf x}_{vc}$ is achieved from a weighted average among all virtual vehicles whose corresponding landmarks are observed right now as:
$$
{\bf x}_{vc}=(\sum_{i\in O}\Sigma_{vi}^{-1})^{-1} \sum_{i\in O}(\Sigma_{vi}^{-1}{\bf x}_{vi})
$$  
Here $O$ is the set of landmarks observed by the robot at that moment. Covariance matrices $\Sigma_{vi}$'s are the components related to the states ${\bf x}_{vi}$'s in the covariance matrices ${\bf P}_i$'s from each distributed small scale Kalman filters.
$${\bf P}_i=\begin{bmatrix}
\Sigma_{i}& \Sigma_{ivi}\\
\Sigma_{vii}& \Sigma_{vi}
\end{bmatrix}$$
The weighted average above is the least square result summarizing information from all observations. Since we already have the virtual vehicle estimations at ${\bf x}_{vi}$'s with covariance matrices $\Sigma_{vi}$'s, we can use the virtual vehicles ${\bf x}_{vi}$'s as noisy measurements about true ${\bf x}_{v}$. To summarize information from all virtual vehicles, we want to find the best estimation of ${\bf x}_{v}$ among these measurements to minimize the quadratic error:
$$
e_v=E(\sum_{i\in O}||{\bf x}_{v}-{\bf x}_{vi}||^2)
$$

whose solution is $$
{\bf x}_{vc}=(\sum_{i\in O}\Sigma_{vi}^{-1})^{-1} \sum_{i\in O}(\Sigma_{vi}^{-1}{\bf x}_{vi})
$$
This weighted average result can also be thought of as a Kalman filter for a system with no dynamics, with virtual vehicles corresponding to the measurements.

We can simultaneously feed the consensus result to the whole network as a leader for all ${\bf x}_{vi}$'s, by treating the consensus ${\bf x}_{vc}$ as a virtual measurement that each virtual vehicle ${\bf x}_{vi}$ could observe. 
$${\bf y}_{i2}={\bf x}_{vc}$$ $${\bf H}_{i2}=[{\bf 0}\quad I]$$
Similar to the full Kalman filter we introduced in Chapter 4, we can update the heading state $\beta$ using a separate optimized estimator following $\beta_d$ , which minimizes the quadratic residue error
$$\beta_d=argmin_{\beta_d\in[-\pi,\pi]}({\bf y}^T-{\bf x}_{vc}^T{\bf H}^T)({\bf y-Hx_{vc}})$$
and
$$\dot{\hat{\beta}}=\omega+\gamma_{\beta}(\beta_d-\hat{\beta})$$
For the motion of the vehicle, we also have 
$${\bf u}=\begin{bmatrix}
u\sin\hat{\beta}\\
u\cos\hat{\beta}
\end{bmatrix}$$
We want all virtual vehicles to converge to the consensus because we want information gathered from all landmarks being summarized at the consensus to be able to get distributed back to influence mapping of all landmarks. Thus, update on the vehicle's location is not isolated, but it can provide corrections for landmarks through virtual vehicles. In such case, even though we don't have covariance matrix to correlate different landmarks with each other, the virtual strings of virtual vehicles to the consensus still would be able to linkage different landmarks through virtual vehicles and distributed small scale covariance matrices.

\subsection{Complete algorithm}
In summary, the algorithm we propose here is composed of two levels of computation: the first level uses separate LTV Kalman filters for each single pair of landmark and virtual vehicle, including both the measurements and the following behavior towards the consensus for the virtual vehicle. The second level is to gather information from all virtual vehicles that have their corresponding landmarks under observation. The consensus ${\bf x}_{vc}$ is the best estimation from the weighted average that minimize the square error.
\\
For all landmarks ${\bf x}_i$ and virtual vehicles ${\bf x}_{vi}$:
$$
{\bf y}_{i}=\left[\begin{array}{c} {\bf y_{i1}}\\{\bf y_{i2}}\end{array}\right]\quad \text{and}\quad {\bf H}_{i}=\left[\begin{array}{c} {\bf H}_{i1}\\{\bf H}_{i2}\end{array}\right]
$$
And the LTV Kalman filter for each virtual vehicle is:
$$
\left[\begin{array}{c}\dot{\hat{{\bf x}}}_{i}\\{\dot{\hat{{\bf x}}}_{vi}}\end{array}\right]=\left[\begin{array}{c}0\\{\bf u}\end{array}\right]+{\bf P}_i{\bf H}_i^T{\bf R}^{-1}({\bf y}_i-{\bf H}_i\left[\begin{array}{c}{\hat{{\bf x}}}_{i}\\{{\hat{{\bf x}}}_{vi}}\end{array}\right])$$
$$
\dot{{\bf P}}_i={\bf Q-P}_i{\bf H}_{i}^T{\bf R}^{-1}{\bf H}_{i}{\bf P}_i$$
$$
{\bf x}_{vc}=(\sum_{i\in O}\Sigma_{vi}^{-1})^{-1} \sum_{i\in O}(\Sigma_{vi}^{-1}{\bf x}_{vi})
$$  
$$\beta_d=argmin_{\beta_d\in[-\pi,\pi]}\Sigma_i({\bf y}_i-{\bf H}_i\left[\begin{array}{c}{\hat{{\bf x}}}_{i}\\{{\hat{{\bf x}}}_{vi}}\end{array}\right])^T({\bf y}_i-{\bf H}_i\left[\begin{array}{c}{\hat{{\bf x}}}_{i}\\{{\hat{{\bf x}}}_{vi}}\end{array}\right])$$
and
$$\dot{\hat{\beta}}=\omega+\gamma_{\beta}(\beta_d-\hat{\beta})$$
$${\bf u}=\begin{bmatrix}
u\sin\hat{\beta}\\
u\cos\hat{\beta}
\end{bmatrix}$$

\subsection{Experiment on Victoria Park benchmarks}
\begin{figure*}[htbp] 
\centering 
\includegraphics[width=1\textwidth]{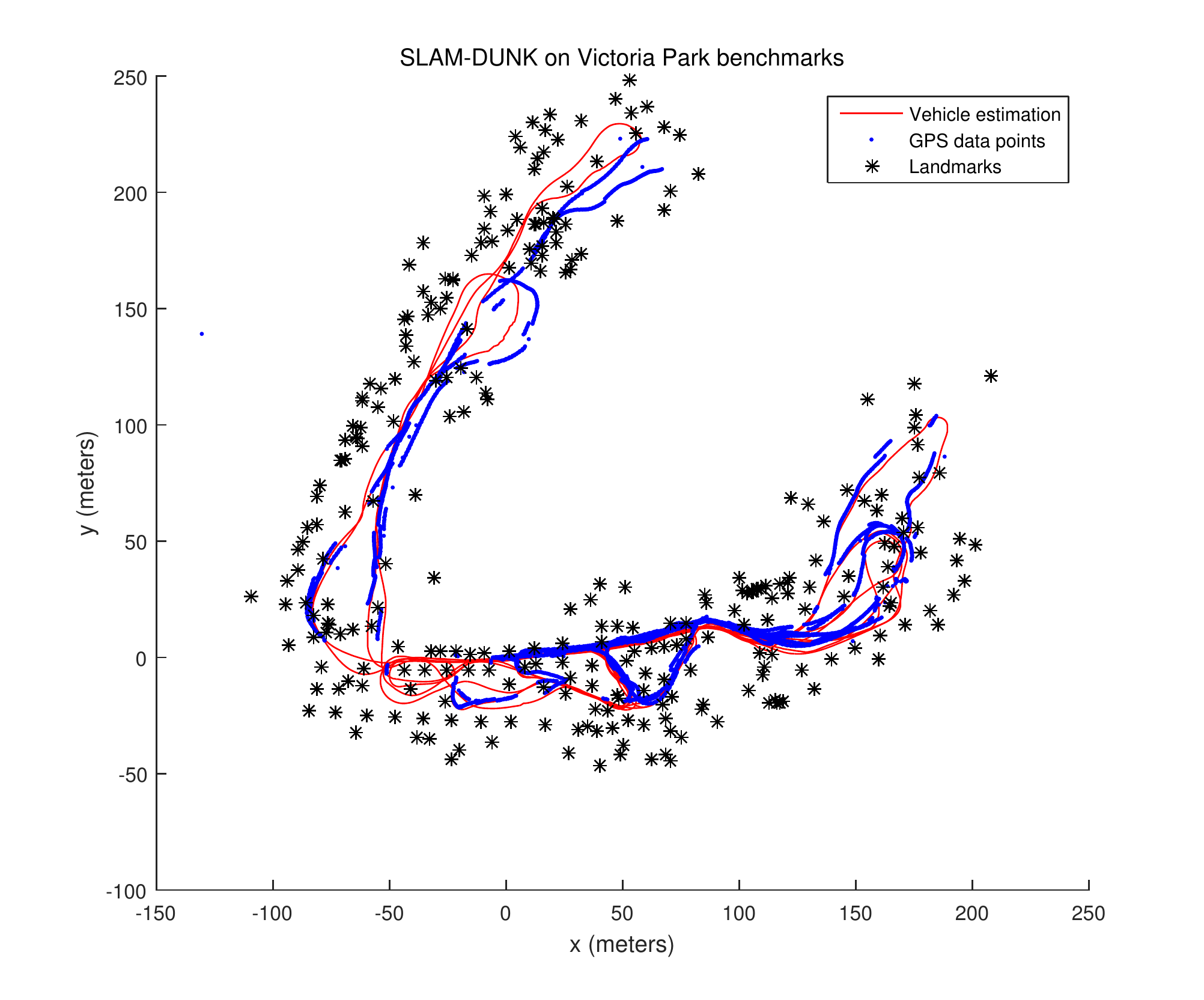}

\caption{ Path and landmarks estimation of full SLAM-DUNK. The thick blue path is the GPS data and the solid red path is the estimated path; the black asterisks are the estimated positions of the landmark.}
\label{virtual}
\end{figure*}

Similar to the full LTV Kalman filter, we applied our algorithm to Sydney Victoria Park dataset. Our algorithm still achieves satisfying result shown in Fig. \ref{virtual} comparing favorably to benchmark result of Unscented FastSLAM \cite{uslam}. Full simulation video of the Victoria Park dataset is provided at \url{https://vimeo.com/173641447}. Covariance ellipses are also included in the simulation.

\subsection{Remarks}
Under certain situations, there would be some special cases for the proposed algorithm. When the landmark $i$ is not observed by the vehicle, the observation parts of the first level estimator would be dropped, that means ${\bf y}_{i1}$ and ${\bf H}_{i1}$ would not be included. When the vehicle sees no landmark at any moment, ${\bf y}_{i2}$ and ${\bf H}_{i2}$ would be dropped, because there would be no way to achieve ${\bf x}_{vc}$ from weight average.
\\
When the vehicle sees a new landmark for the first time, the corresponding virtual vehicle is initialized at the location of current best estimation, which can be computed from weight average among all virtual vehicles as:
$${\bf x}_{vi0}=(\sum_{j\in all}\Sigma_{vj}^{-1})^{-1} \sum_{j\in all}(\Sigma_{vj}^{-1}{\bf x}_{vj})$$
Furthermore, data association to match the observed landmarks with the ones in the memory can be simply carried by matching the measurements with the saved pairs of landmarks and virtual vehicles.
\\
Note that for each pair of landmark and virtual vehicle, we design an LTV Kalman filter specifically. That means we would have $n$ filters in total, where $n$ is the total number of landmarks. For each filter, it would have two states to estimate. The total computation complexity would be $O(n)$, which would be comparable to FastSLAM with only two particles, while FastSLAM with only two particles would sacrifice on performance significantly. Moreover, for filters whose landmarks are not observed, they only have the behavior of following, so the computation is even lighter. 
\\
The whole idea of the proposed algorithm is to break the full LTV Kalman filter containing both the landmarks and the vehicle states into $n$ small estimators to get the best estimation locally and one optimization of least squares to achieve the best estimation on consensus. We successfully decouple the landmarks, in addition, since each single estimator still follows the same structure as the full ones proposed before, contraction analysis that is identical to the full LTV Kalman filters can be exploited to ensure the estimations would finally converge to the noise-free true states.
\\
Despite the authors' lack of enthusiasm for acronyms, it is hard to resist calling the final algorithm SLAM-DUNK \footnote{We thank Geoffrey Hinton for the suggestion.}, 
for simultaneous location and mapping using distributed unlinearized networked Kalman-filtering.

\section{Distributed Multi-robot Cooperative SLAM without Prior Global Information}

In this chapter we inherit the proposed distributed algorithm SLAM-DUNK and further extend the discussion to algorithms for multi-robot cooperative SLAM. Multi-robot cooperative SLAM has become more and more an important and meaningful problem to study. As light robot systems like drones or ground vehicles become more and more inexpensive and accessible, research on how to utilize distributed computation power and swarms of robots to benefit performance of localization, mapping and exploration would give insights to future developments. In this chapter, we propose algorithms for cooperative SLAM in different scenarios, an all-observable setting, a case where robots have incomplete observations and finally a robot-only case. Discussion in this chapter has also drawn inspirations from quorum sensing, a phenomenon of group behavior coordination in nature. The focus of this chapter is on how to make sure robots starting from different poses and positions could share information and all converge to a shared global map for collaborative exploration.

\subsection{Basic assumptions in this chapter}

Before we introduce the proposed algorithms for cooperative SLAM, let us introduce the basic assumptions for defining the setting of the problem.

\begin{itemize}

\item A group of $M$ independent robots move in a 2D space with $N$ features. Each robot moves on their own with their respective kinematics and dynamics. They have proprioceptive sensing devices to measure the self motion of the robot. 

\item The robots also carry exteroceptive sensing devices to monitor and observe the environment for localization features such as landmarks. Such devices could be cameras, lidars, sonars, etc. to measure different information like bearing or range, or even more.

\item Each robot also has the capability to measure the other nearby robots with relative measurements. Such measurements are not constrained to the traditional bearing and range, but also relative pose difference such as heading, which can be analyzed from camera images.

\item Agent-to-agent information communication is not necessarily required. However, we assume the robots could communicate with a central medium, submitting their local map of landmarks and their velocities. The central medium is able to feedback each agent with mingled information for them to synchronize with the medium and indirectly to each other.

\item We also assume that data association is not of our major concern in discussion here. As visual features become more and more accessible, identifying features from one to another also becomes easier.

\item We do not require the robots to know their initial global positions and poses. Our algorithms are designed to deal with the differences between coordinates systems of robots automatically.  

\end{itemize}

In comparison to most of the existing works that solve the problem with one complete state vector including all landmarks and all robots, we focus on developing a general framework of distributed Kalman filters, where each robot keeps its own map of environment, and tunes the map based on feedback from the medium to finally converge with all other maps to reach a consensus.

\subsection{Basic idea of null space}
Let us take another look at the algorithm we proposed in Chapter 5:
$$
\left[\begin{array}{c}\dot{\hat{{\bf x}}}_{i}\\{\dot{\hat{{\bf x}}}_{vi}}\end{array}\right]=\left[\begin{array}{c}0\\{\bf u}\end{array}\right]+{\bf P}_i{\bf H}_i^T{\bf R}^{-1}({\bf y}_i-{\bf H}_i\left[\begin{array}{c}{\hat{{\bf x}}}_{i}\\{{\hat{{\bf x}}}_{vi}}\end{array}\right])$$
$$
\dot{{\bf P}}_i={\bf Q-P}_i{\bf H}_{i}^T{\bf R}^{-1}{\bf H}_{i}{\bf P}_i$$
The basic idea of all algorithms discussed in this chapter is that there is a null space problem, or sometimes described as observability problem in the SLAM model. For the map consisted of landmarks and the vehicle, if there is no global information available, it is free of translation and rotation.
$$
\left[\begin{array}{c}\dot{\hat{{\bf x}}}_{i}\\{\dot{\hat{{\bf x}}}_{vi}}\end{array}\right]=({\bf v}+\Omega \left[\begin{array}{c}{\hat{{\bf x}}}_{i}\\{{\hat{{\bf x}}}_{vi}}\end{array}\right] )+ \left[\begin{array}{c}0\\{\bf u}\end{array}\right]+{\bf P}_i{\bf H}_i^T{\bf R}^{-1}({\bf y}_i-{\bf H}_i\left[\begin{array}{c}{\hat{{\bf x}}}_{i}\\{{\hat{{\bf x}}}_{vi}}\end{array}\right])$$
That means as shown in the equation above, we can freely add any translation or rotation term $$({\bf v}+\Omega \left[\begin{array}{c}{\hat{{\bf x}}}_{i}\\{{\hat{{\bf x}}}_{vi}}\end{array}\right] )$$ to the equation as long as the inputs $\bf v$ and $\Omega$ are the same for all landmarks belong to the map of the same robot. We call such freely added translation and rotation terms as null space terms. Such action will have no impact to the map and localization of the robot, since all relative constraints between landmarks stay the same because for $$\forall i,j\ \text{and}\ {\bf v},\ ||{\bf x}_i+{\bf v}-({\bf x}_j+{\bf v})||= ||{\bf x}_i-{\bf x}_j|| $$ and $$\forall i,j\ \text{and}\ {\bf R},\ ||{\bf R}_{rotation}{\bf x}_i-{\bf R}_{rotation}{\bf x}_j||= ||{\bf x}_i-{\bf x}_j|| $$
where ${\bf R}_{rotation}$ is a rotational matrix. In such case, all relative constraints are preserved, and the map after such transformations has not been influenced at all.

Therefore, the problem to consider is how to make use of this null space and get maps from different robots to converge to a unified map with the same coordinate system. We will introduce in the following sections on detailed algorithm design.

Here we introduce the terms we use to define the environment and the model. Assume there are $M$ robots and $N$ landmarks in the environment, ${\bf x}_{ik}$ is the position of landmark $k$ in the $i$th robot's coordinates. ${\bf x}_{iv}$ and $\beta_i$ are the position and heading of the $i$th robot. ${\bf v}_i$ is the translation velocity in null space and $\Omega_i=\begin{bmatrix}
0&-\omega\\
\omega&0\\
\end{bmatrix}$ is the corrective angular velocity in null space. For the other terms, we inherit them from Chapter 5. So the system turns to $$
\left[\begin{array}{c}\dot{\hat{{\bf x}}}_{ik}\\{\dot{\hat{{\bf x}}}_{iv}}\end{array}\right]=({\bf v}_i+\Omega_i \left[\begin{array}{c}{\hat{{\bf x}}}_{ik}\\{{\hat{{\bf x}}}_{iv}}\end{array}\right] )+ \left[\begin{array}{c}0\\{\bf u}_i\end{array}\right]+{\bf P}_{ik}{\bf H}_{ik}^T{\bf R}^{-1}({\bf y}_{ik}-{\bf H}_{ik}\left[\begin{array}{c}{\hat{{\bf x}}}_{ik}\\{{\hat{{\bf x}}}_{iv}}\end{array}\right])$$ 
And for the covariances:
$$
\dot{{\bf P}}_{ik}={\bf Q-P}_{ik}{\bf H}_{ik}^T{\bf R}^{-1}{\bf H}_{ik}{\bf P}_{ik}$$

\subsection{Cooperative SLAM with full information}

In this section we introduce the algorithm for the case when all robots could observe all landmarks all them time, which we call the scenario ``all-know-all''. In such a setting, we assume that each robot has measurements of all landmarks, so that every robot has the same level of information, and the problem turns to how to merge all information in a shared coordinate system. 

As we have defined earlier, ${\bf x}_{ik}$ is the position of landmark $k$ in the $i$th robot's coordinates. We denote the virtual center of all landmarks observed in the $i$th robot's coordinate system as $${\bf x}_{ic}=\frac{1}{N}\Sigma_k{\bf x}_{ik}$$ For all coordinate systems to converged to a consensus, the first step is the have their average center to converge to each other and then finally to the same point. That means, for the final result $$\forall i,j,\ ||{\bf x}_{ic}-{\bf x}_{jc}||=0$$

We can rewrite the null space term's rotation part to have the rotation center at ${\bf x}_{ic}$: $${\bf v}_i+\Omega_i \left[\begin{array}{c}{{{\bf x}}}_{ik}-{\bf x}_{ic}\\{{{{\bf x}}}_{ikv}}-{\bf x}_{ic}\end{array}\right]$$
In this case, the translation and rotation parts of the null space terms are entirely independent from each other. When we think about how to get all ${\bf x}_{ic}$'s to converge to each other, only the choice of ${\bf v}_i$ as an input matters, and the rotation parts have no influence here. We can easily choose ${\bf v}_i$ to minimize
$$e_c=\Sigma_j||{\bf x}_{ic}-{\bf x}_{jc}||^2$$
To minimize the center error, we can choose 
$${\bf v}_i=\gamma_{vi}\Sigma_j({\bf x}_{jc}-{\bf x}_{ic})$$
To further borrow ideas from quorum sensing, we have no need to take computation for each pair, but just obtain a shared medium as the consensus of the center of each robot-landmark system as $${\bf x}_{cc}=\frac{1}{M}\Sigma_i{\bf x}_{ic}$$
And we can change the inputs of ${\bf v}_i$'s to be $${\bf v}_i=\gamma_{vi}({\bf x}_{cc}-{\bf x}_{ic})$$

Simply making sure the centers of all coordinate systems converge to the same consensus does not guarantee all coordinate systems to be the same, because there still leave differences among headings. As we use the rotation part in the null space term $\Omega_i \left[\begin{array}{c}{\bf x}_{ik}-{\bf x}_{ic}\\{\bf x}_{ikv}-{\bf x}_{ic}\end{array}\right]$ to help the systems to converge to a shared heading, we bring up a new metric to optimize, as the ideal system should have $$\forall i,j\ \text{and}\ k, ||{\bf x}_{ik}-{\bf x}_{ic}-({\bf x}_{jk}-{\bf x}_{jc})||=0$$
That means we should minimize the so defined heading error 
$$e_h=\Sigma_k\Sigma_j||{\bf x}_{ik}-{\bf x}_{ic}-({\bf x}_{jk}-{\bf x}_{jc})||^2$$
with input $\Omega_i=\begin{bmatrix}
0&-\omega_i\\
\omega_i&0
\end{bmatrix}$ 

To find the proper input $\omega_i$, we analyze as:
\begin{eqnarray}
\frac{d}{dt}e_h&=&\Sigma_k\Sigma_j(\dot{\bf x}_{ik}-\dot{\bf x}_{ic})^T[{\bf x}_{ik}-{\bf x}_{ic}-({\bf x}_{jk}-{\bf x}_{jc})]\nonumber\\
&=&\Sigma_k\Sigma_j({\bf x}_{ik}-{\bf x}_{ic})^T\begin{bmatrix}
0&\omega_i\\
-\omega_i&0
\end{bmatrix}[{\bf x}_{ik}-{\bf x}_{ic}-({\bf x}_{jk}-{\bf x}_{jc})]\nonumber\\
&=&\omega\Sigma_k\Sigma_j({\bf x}_{ik}-{\bf x}_{ic})^T\begin{bmatrix}
0&1\\
-1&0
\end{bmatrix}({\bf x}_{jc}-{\bf x}_{jk})\nonumber\\
\end{eqnarray}  
Here we are inspired by quorum sensing again, to utilize another medium variable as 
$${\bf x}_{ck}=\frac{1}{M}\Sigma_i{\bf x}_{ik}$$ This medium variable is the temporary average of any feature $k$ among all robot-landmarks coordinate systems. Since ${\bf x}_{cc}=\frac{1}{M}\Sigma_i{\bf x}_{ic}$, we can have 
\begin{eqnarray}
\frac{d}{dt}e_h&=&\omega\Sigma_k\Sigma_j({\bf x}_{ik}-{\bf x}_{ic})^T\begin{bmatrix}
0&1\\
-1&0
\end{bmatrix}({\bf x}_{jc}-{\bf x}_{jk})\nonumber\\
&=&\omega M \Sigma_k({\bf x}_{ik}-{\bf x}_{ic})^T\begin{bmatrix}
0&1\\
-1&0
\end{bmatrix}({\bf x}_{cc}-{\bf x}_{ck})\nonumber\\
\end{eqnarray} 
Since $\Sigma_k({\bf x}_{ik}-{\bf x}_{ic})=0$
\begin{eqnarray}
\frac{d}{dt}e_h&=&\omega M \Sigma_k({\bf x}_{ik}-{\bf x}_{ic})^T\begin{bmatrix}
0&1\\
-1&0
\end{bmatrix}({\bf x}_{cc}-{\bf x}_{ck})\nonumber\\
&=&-\omega M \Sigma_k({\bf x}_{ik}-{\bf x}_{ic})^T\begin{bmatrix}
0&1\\
-1&0
\end{bmatrix}{\bf x}_{ck}\nonumber\\
\end{eqnarray} 
To make sure $\frac{d}{dt}e_h\leq 0$, and that $e_h$ keeps getting reduced, we can choose inputs $\omega_i$'s to be
$$\omega_i=\gamma_{\omega i}\Sigma_k({\bf x}_{ik}-{\bf x}_{ic})^T\begin{bmatrix}
0&1\\
-1&0
\end{bmatrix}{\bf x}_{ck}$$
To summarize, we can keep reducing center differences $e_c$ and heading differences $e_h$ as we utilize medium variables ${\bf x}_{ic}$'s, ${\bf x}_{cc}$'s, ${\bf x}_{ck}$'s and implement the inputs ${\bf v}_i$'s and $\omega_i$'s as
$${\bf v}_i=\gamma_{vi}({\bf x}_{cc}-{\bf x}_{ic})$$
$$\omega_i=\gamma_{\omega i}\Sigma_k({\bf x}_{ik}-{\bf x}_{ic})^T\begin{bmatrix}
0&1\\
-1&0
\end{bmatrix}{\bf x}_{ck}$$
And use them as inputs to null space terms in the SLAM-DUNK algorithms as 
$$
\left[\begin{array}{c}\dot{\hat{{\bf x}}}_{ik}\\{\dot{\hat{{\bf x}}}_{ikv}}\end{array}\right]=({\bf v}_i+\Omega_i \left[\begin{array}{c}{\hat{{\bf x}}}_{ik}-{\hat{{\bf x}}}_{ic}\\{{\hat{{\bf x}}}_{ikv}}-{\hat{{\bf x}}}_{ic}\end{array}\right] )+ \left[\begin{array}{c}0\\{\bf u}_i\end{array}\right]+{\bf P}_{ik}{\bf H}_{ik}^T{\bf R}^{-1}({\bf y}_{ik}-{\bf H}_{ik}\left[\begin{array}{c}{\hat{{\bf x}}}_{ik}\\{{\hat{{\bf x}}}_{ikv}}\end{array}\right])$$ 
$$
\dot{{\bf P}}_{ik}={\bf Q}_i-{\bf P}_{ik}{\bf H}_{ik}^T{\bf R}^{-1}{\bf H}_{ik}{\bf P}_{ik}$$
where
$$
{\bf x}_{ivc}=(\sum_{k\in O}\Sigma_{ivk}^{-1})^{-1} \sum_{k\in O}(\Sigma_{ivk}^{-1}{\bf x}_{ivk})
$$  
$$\beta_{di}=argmin_{\beta_d\in[-\pi,\pi]}\Sigma_k({\bf y}_{ik}-{\bf H}_{ik}\left[\begin{array}{c}{\hat{{\bf x}}}_{ik}\\{{\hat{{\bf x}}}_{ikv}}\end{array}\right])^T({\bf y}_{ik}-{\bf H}_{ik}\left[\begin{array}{c}{\hat{{\bf x}}}_{ik}\\{{\hat{{\bf x}}}_{ikv}}\end{array}\right])$$
$$\dot{\hat{\beta}}_i=\omega_{im}+\omega_i+\gamma_{\beta_i}(\beta_{di}-\hat{\beta}_i)$$
$${\bf u}_i=\begin{bmatrix}
u_i\sin\hat{\beta}_i\\
u_i\cos\hat{\beta}_i
\end{bmatrix}$$ 
As we stated earlier, the null space terms has no influence over the main algorithm, so the contraction property is preserved and the true locations of both the landmarks and the robots in the shared global coordinate system consist a particular solution to every robot's distributed filters. As all the states finally converge to be static, so will ${\bf v}_i$'s and $\omega_i$'s converge and stay zero, which ensures $e_h$ and $e_c$ to reduce to zero. Thus, benefiting from both the filters and the null terms, all robots will converge to the same coordinate system regardless of their initial states.

\subsubsection{Simulation results}
\begin{figure}[htbp] 
\centering 
\centerline{\includegraphics[width=1\textwidth]{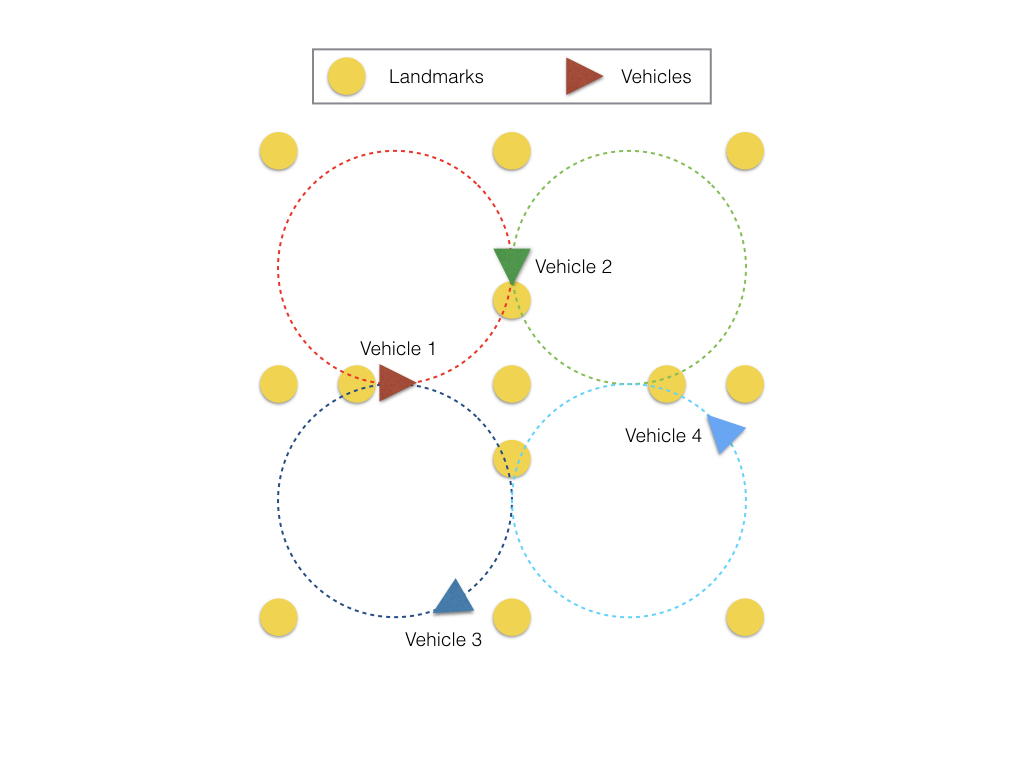}}
\caption{Simulation environment for cooperative SLAM with full information. We have 13 landmarks as circles and 4 vehicle as triangles}
\label{group}
\end{figure}

Here we provide simulation results for the proposed algorithm on cooperative SLAM with full information. As shown in Fig. \ref{group}, we have 13 landmarks and 4 vehicles in a synthetic environment. The thirteen landmarks are located at [-30, 30], [0, 30], [30, 30], [-30, 0], [0, 0], [30, 0], [-30, -30], [0, -30], [30, -30], [0, 10], [0, -10], [-20, 0], and [20, 0], and the four vehicles are located in the four quadrants among the landmarks. The four vehicles are circling respectively around the centers $c_1=[-15,\ 15]$, $c_2=[15,\ 15]$, $c_3=[-15,\ -15]$ and $c_4=[15,\ -15]$ with radius $15$m. Their angular velocities are different as $\omega_{m1}=1\ rad/s$, $\omega_{m2}=1.5\ rad/s$, $\omega_{m3}=-1\ rad/s$, and $\omega_{m4}=0.5\ rad/s$. All vehicles start from different initial positions and different initial headings as 
$${\bf x}_{10}=[-15,\ 0]\ \ \text{and}\ \ \beta_{10}=0$$   
$${\bf x}_{20}=[0,\ 15]\ \ \text{and}\ \ \beta_{20}=\frac{3}{2}\pi$$
$${\bf x}_{30}=[-7.5,\ -7.5-7.5\sqrt{3}]\ \ \text{and}\ \ \beta_{30}=\frac{7}{6}\pi$$
$${\bf x}_{40}=[15+7.5\sqrt{2},\ -15+7.5\sqrt{2}]\ \ \text{and}\ \ \beta_{40}=\frac{3}{4}\pi$$

But since all vehicles have no prior global information about their starting positions and starting headings, they all start in their own local coordinates at ${\bf x}_0=[0, 0]$ with heading $\beta_0=0$. Then we implement our proposed algorithm for cooperative SLAM with full information on the synthetic simulation environment. As we can see from the full video on \url{https://vimeo.com/193489754}, even with no prior global information and simply using their local coordinates as starting points, coordinate systems from different vehicles shift and rotate to converge to each other. The red, green, blue and cyan landmarks and dashed lines of vehicle trajectories correspond respectively to estimations from vehicles 1, 2, 3 and 4. We can see that after roughly 10 seconds, landmark estimations from different vehicles converge to reach consensus, and trajectories of vehicles also converge to the circles as they are expected to be. Keep in mind that since we have no global information available, the achieved consensus result is a rotated and shifted transformation from the truth. As long as relative constraints remain intact, we can consider the algorithm to achieve a true map from the consensus.

\begin{figure}[htbp] 
\centering 
\includegraphics[width=\textwidth]{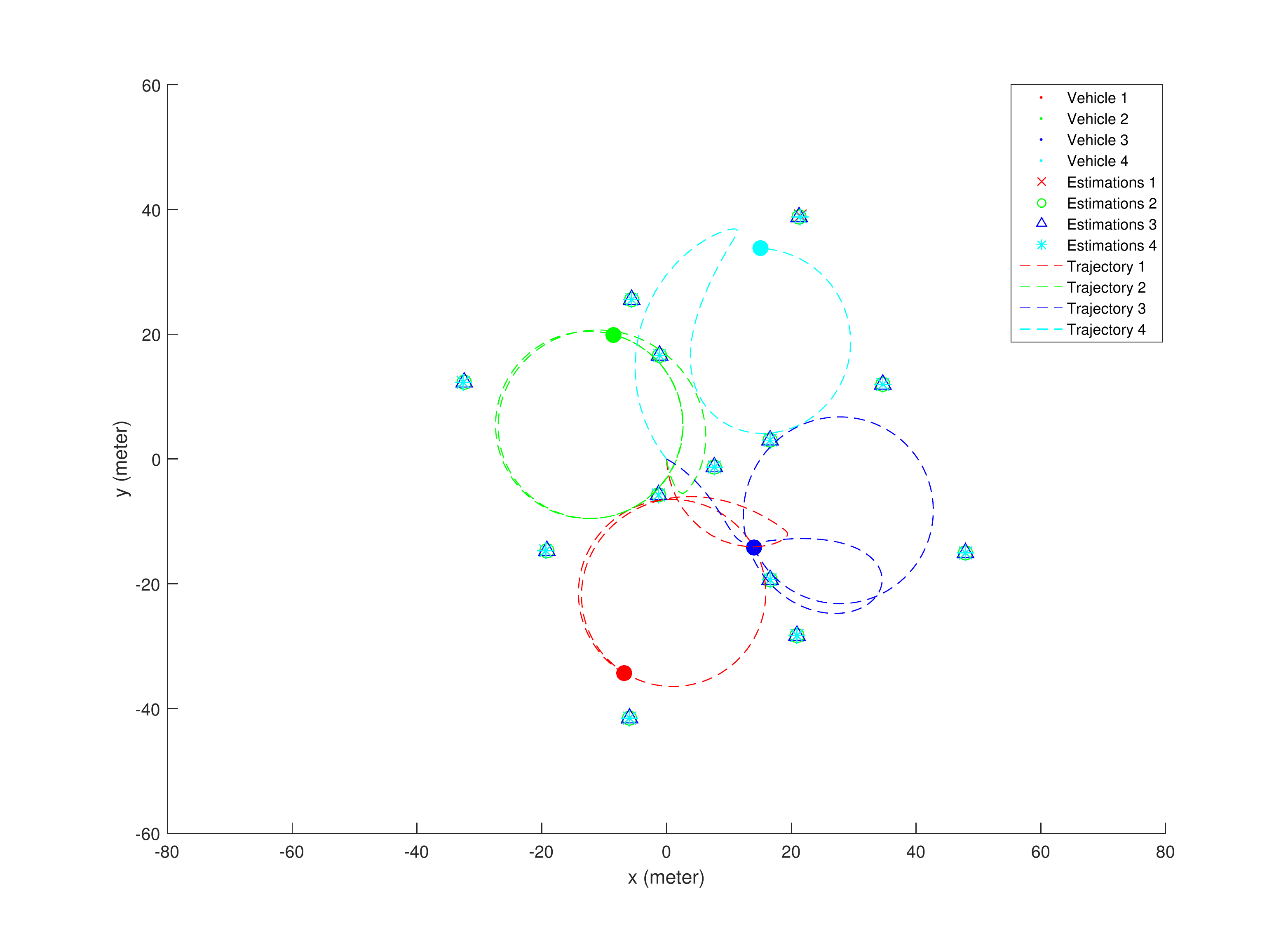}
\caption{Simulation results for cooperative SLAM with full information. The red, green, blue and cyan landmarks and dashed lines of vehicle trajectories correspond respectively to estimations from vehicles 1, 2, 3 and 4}
\label{group1f}
\end{figure}

\subsection{Cooperative SLAM with partial information}

In the section above we discussed about algorithm to be used when all robots keep observing all landmarks. However, that is not a usual case because oftentimes, robots could not see all landmarks in the map due to distance, occlusion, feature selection and other factors. In addition, one of the major advantages for cooperative SLAM is the capability to gather partial information from each robot and stitch them together for complete and global information. For example, one can use a group of robots to explore an unknown area and achieve information in a more accurate and much faster way as they can be sent to different directions, heights, and even be equipped with different sensors. In this section we propose a general  algorithm that could perform cooperative SLAM among robots with potentially partial information.

\subsubsection{Nearest neighbor as feature}
One problem in cooperative SLAM in different coordinates is that global information is not accessible, and how to build connections between different coordinate systems becomes a problem. As we suggested in earlier sections, relative positions between landmarks $||{\bf x}_i-{\bf x}_j||$ is invariant in different coordinates, and that extends to $||{\bf x}_i-{\bf x}_c||$, which is used in the earlier section. Yet, when we start to looking into relative positions between landmarks, the complexity turns to $O(n^2)$. Since only some of the landmarks can be observed by one robot, it would be hard to use a common metric for all different robots. That is also why we used $||{\bf x}_i-{\bf x}_c||$ in the last section. Thus, we also need to develop an algorithm that keeps the complexity to be $O(n)$, while making sure that the metric is invariant to translations and rotations. That is the reason in this section we use nearest neighbors as feature vectors.

The general process is relatively straightforward. For each robot $i$, and any landmark $k$ observed by the robot, the robot finds the nearest landmark $k'$ that is closest to $k$, which means
$$x_{ik'}=argmin_{k'\neq k\in O_i}||{\bf x}_{ik}-{\bf x}_{ik'}||$$
where $O_i$ is the set of features that robot $i$ observes. Then robot $i$ report the identity of closest neighbor $k'$ along with the observed vector ${\bf a}_{ik}={\bf x}_{ik}-{\bf x}_{ik'}$ to the central coordinator. The central coordinator collects information from all robots that can observe landmark $k$ and compare the results $||{\bf a}_{ik}||$'s to determine the true nearest neighbor of landmark $k$ as $k*$. Each robot will receive the feedback to confirm whether $k'=k*$. If it does not match, then robot $i$ would not have a nearest-neighbor feature for landmark $k$, but if it does match as $k'=k*$, robot $i$ will have the observation of the nearest-neighbor feature for landmark $k$ as 
$${\bf a}_{ik}={\bf x}_{ik}-{\bf x}_{ik'}$$ 
In such case, the shared map that all robots will converge to is a map of unidirectional vectors ${\bf a}_{ik}$'s and the number of these nearest-neighbor feature vectors would be same as number of landmarks $N$. Since nearest neighbor is translation and rotation invariant, it provides a shared anchor for all robots.

\subsubsection{Algorithm for cooperative SLAM with partial information}
After we defined the nearest-neighbor feature vectors, the algorithm part becomes much more straightforward and the structure is similar to what we have in the proposal for full information.

First we have a similar same structure here:
$$
\left[\begin{array}{c}\dot{\hat{{\bf x}}}_{ik}\\{\dot{\hat{{\bf x}}}_{iv}}\end{array}\right]=({\bf v}_i+\Omega_i \left[\begin{array}{c}{\hat{{\bf x}}}_{ik}\\{{\hat{{\bf x}}}_{iv}}\end{array}\right] )+ \left[\begin{array}{c}0\\{\bf u}_i\end{array}\right]+{\bf P}_{ik}{\bf H}_{ik}^T{\bf R}^{-1}({\bf y}_{ik}-{\bf H}_{ik}\left[\begin{array}{c}{\hat{{\bf x}}}_{ik}\\{{\hat{{\bf x}}}_{iv}}\end{array}\right])$$ 
and same as before we use ${\bf v}_i$ and $\omega_i$ as inputs to help different coordinate systems to converge to a consensus.

Since we are not guaranteed that all landmarks can be observed, there is no way to calculate ${\bf x}_{ic}$ and ${\bf x}_{cc}$, so we use the other medium variable
$${\bf x}_{ck}=\frac{1}{N_k}\Sigma_{i\in O'_k}{\bf x}_{ik}$$
here $O'_k$ is the set of robots who can observe landmark $k$ and $N_k$ is the number of robots who can observe landmark. We do not require complete observation of the landmark from all robots, with partial information being sufficient.

In that case we design
$${\bf v}_i=\gamma_{vi}\Sigma_k({\bf x}_{ck}-{\bf x}_{ik})$$
For the rotational part we try to rotate to match up ${\bf a}_{ik}$'s, as they are translation invariant. To minimize heading error
$$e_h=\Sigma_k\Sigma_j||{\bf a}_{ik}-{\bf a}_{jk}||^2$$
Since
\begin{eqnarray}
\frac{d}{dt}e_h&=&\Sigma_k\Sigma_j\dot{\bf a}_{ik}^T({\bf a}_{ik}-{\bf a}_{jk})\nonumber\\
&=&\Sigma_k\Sigma_j{\bf a}_{ik}^T\begin{bmatrix}
0&\omega_i\\
-\omega_i&0
\end{bmatrix}({\bf a}_{ik}-{\bf a}_{jk})\nonumber\\
&=&-\omega_i\Sigma_k\Sigma_j{\bf a}_{ik}^T\begin{bmatrix}
0&1\\
-1&0
\end{bmatrix}{\bf a}_{jk}\nonumber\\
\end{eqnarray}
Here we introduce another medium variable ${\bf c}_k$ as
$${\bf c}_k=\frac{1}{N_k*}\Sigma_{i\in O'_k*}{\bf a}_{ik}$$
where $O'_k*$ is the set of robots that can observe the nearest-neighbor feature ${\bf a}_{ik}$ and $N_k*$ is the number of robots who can do that. Thus we have
\begin{eqnarray}
\frac{d}{dt}e_h&=&-\omega_i\Sigma_k\Sigma_j{\bf a}_{ik}^T\begin{bmatrix}
0&1\\
-1&0
\end{bmatrix}{\bf a}_{jk}\nonumber\\
&=&-\omega_i\Sigma_k{\bf a}_{ik}^T\begin{bmatrix}
0&1\\
-1&0
\end{bmatrix}{\bf c}_{k}
\end{eqnarray}
and to make sure $\frac{d}{dt}e_h\leq 0$ we can choose to have $\omega_i$ as 
$$\omega_i=\gamma_{\omega_i}\Sigma_k{\bf a}_{ik}^T\begin{bmatrix}
0&1\\
-1&0
\end{bmatrix}{\bf c}_{k}$$

To summarize, for cooperative SLAM with partial information, we can add a null space term as inputs to the SLAM-DUNK algorithms as 
$$
\left[\begin{array}{c}\dot{\hat{{\bf x}}}_{ik}\\{\dot{\hat{{\bf x}}}_{ikv}}\end{array}\right]=({\bf v}_i+\Omega_i \left[\begin{array}{c}{\hat{{\bf x}}}_{ik}\\{\hat{{\bf x}}}_{ikv}\end{array}\right] )+ \left[\begin{array}{c}0\\{\bf u}_i\end{array}\right]+{\bf P}_{ik}{\bf H}_{ik}^T{\bf R}^{-1}({\bf y}_{ik}-{\bf H}_{ik}\left[\begin{array}{c}{\hat{{\bf x}}}_{ik}\\{\hat{{\bf x}}}_{ikv}\end{array}\right])$$
$$
\dot{{\bf P}}_{ik}={\bf Q}_i-{\bf P}_{ik}{\bf H}_{ik}^T{\bf R}^{-1}{\bf H}_{ik}{\bf P}_{ik}$$
$$
{\bf x}_{ivc}=(\sum_{k\in O}\Sigma_{ivk}^{-1})^{-1} \sum_{k\in O}(\Sigma_{ivk}^{-1}{\bf x}_{ivk})
$$  
$$\beta_{di}=argmin_{\beta_d\in[-\pi,\pi]}\Sigma_k({\bf y}_{ik}-{\bf H}_{ik}\left[\begin{array}{c}{\hat{{\bf x}}}_{ik}\\{{\hat{{\bf x}}}_{ikv}}\end{array}\right])^T({\bf y}_{ik}-{\bf H}_{ik}\left[\begin{array}{c}{\hat{{\bf x}}}_{ik}\\{{\hat{{\bf x}}}_{ikv}}\end{array}\right])$$
$$\dot{\hat{\beta}}_i=\omega_{im}+\omega_i+\gamma_{\beta_i}(\beta_{di}-\hat{\beta}_i)$$
$${\bf u}_i=\begin{bmatrix}
u_i\sin\hat{\beta}_i\\
u_i\cos\hat{\beta}_i
\end{bmatrix}$$ 
where
$${\bf v}_i=\gamma_{vi}\Sigma_k({\bf x}_{ck}-{\bf x}_{ik})$$
and
$$\omega_i=\gamma_{\omega_i}\Sigma_k{\bf a}_{ik}^T\begin{bmatrix}
0&1\\
-1&0
\end{bmatrix}{\bf c}_{k}$$
For definition
$${\bf x}_{ck}=\frac{1}{N_k}\Sigma_{i\in O'_k}{\bf x}_{ik}$$
and
$${\bf c}_k=\frac{1}{N_k*}\Sigma_{i\in O'_k*}{\bf a}_{ik}$$

Same as we discussed in the case with full information, the null space terms have no influence over the main algorithm. So the contraction property is preserved with the noise-free true locations of both the landmarks and the robots in the shared global coordinate system as a particular solution. As all the states finally converge to be static, so will ${\bf v}_i$'s and $\omega_i$'s converge to and stay at zero, which ensures $e_h$ and $e_c$ to reduce to zero. Thus, all robots will converge to the same coordinate system regardless of their initial states and starting points. 

\subsubsection{Simulation results}
\begin{figure}[htbp] 
\centering 
\centerline{\includegraphics[width=1\textwidth]{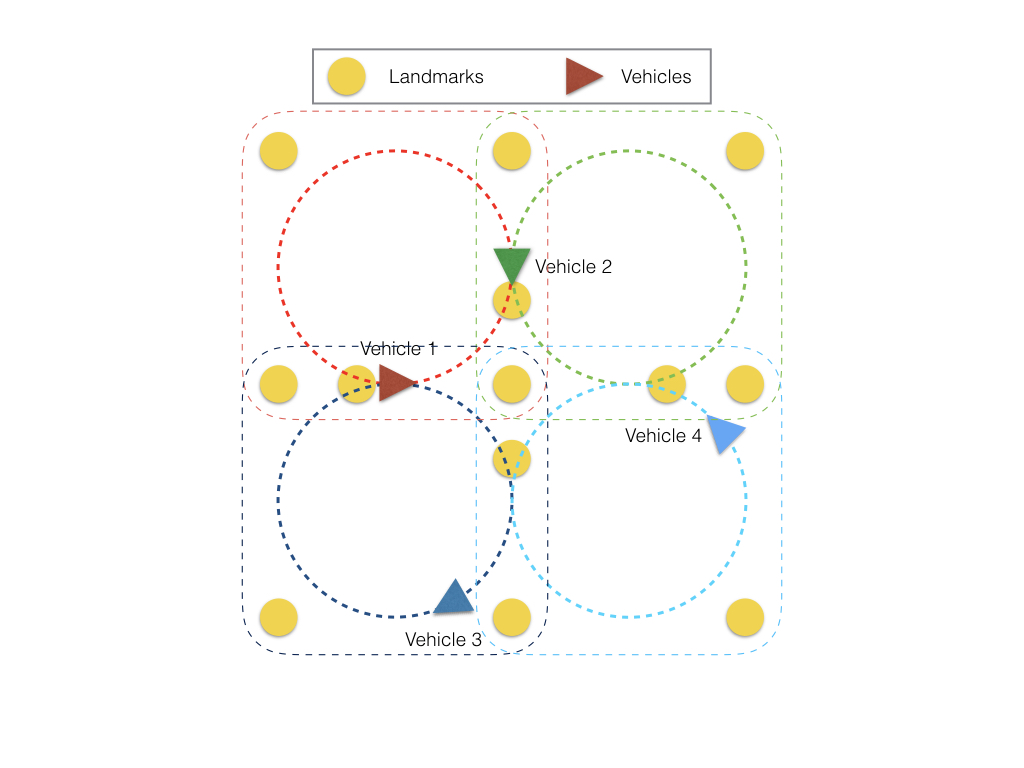}}
\caption{Simulation environment for cooperative SLAM with partial information}
\label{quadrant}
\end{figure}

The simulation environment for cooperative SLAM with partial information is almost identical to the one we proposed in the section with full information, the only difference being that now the robots cannot observe all landmarks. They can only see the landmarks in their respective quadrants while some of them can be observed by multiple vehicles, as shown in Fig. \ref{quadrant}. We use the same initial conditions for the simulation and implement algorithm for cooperative SLAM with partial information. 

As we show in the full video on \url{https://vimeo.com/193489764}, with no prior global information and only partial observation of landmarks, coordinate systems from different vehicles shift and rotate to converge to each other. Similarly, the red, green, blue and cyan landmarks and vehicle trajectories correspond respectively to estimations from vehicles 1, 2, 3 and 4. We can see that landmark estimations from different vehicles converge to reach consensus, and trajectories of vehicles also converge to the circles they are expected to be. As stated before, the achieved consensus result is a rotated and shifted transformation from the truth, and we can consider the algorithm to achieve a true map from the consensus.

\begin{figure}[htbp] 
\centering 
\includegraphics[width=\textwidth]{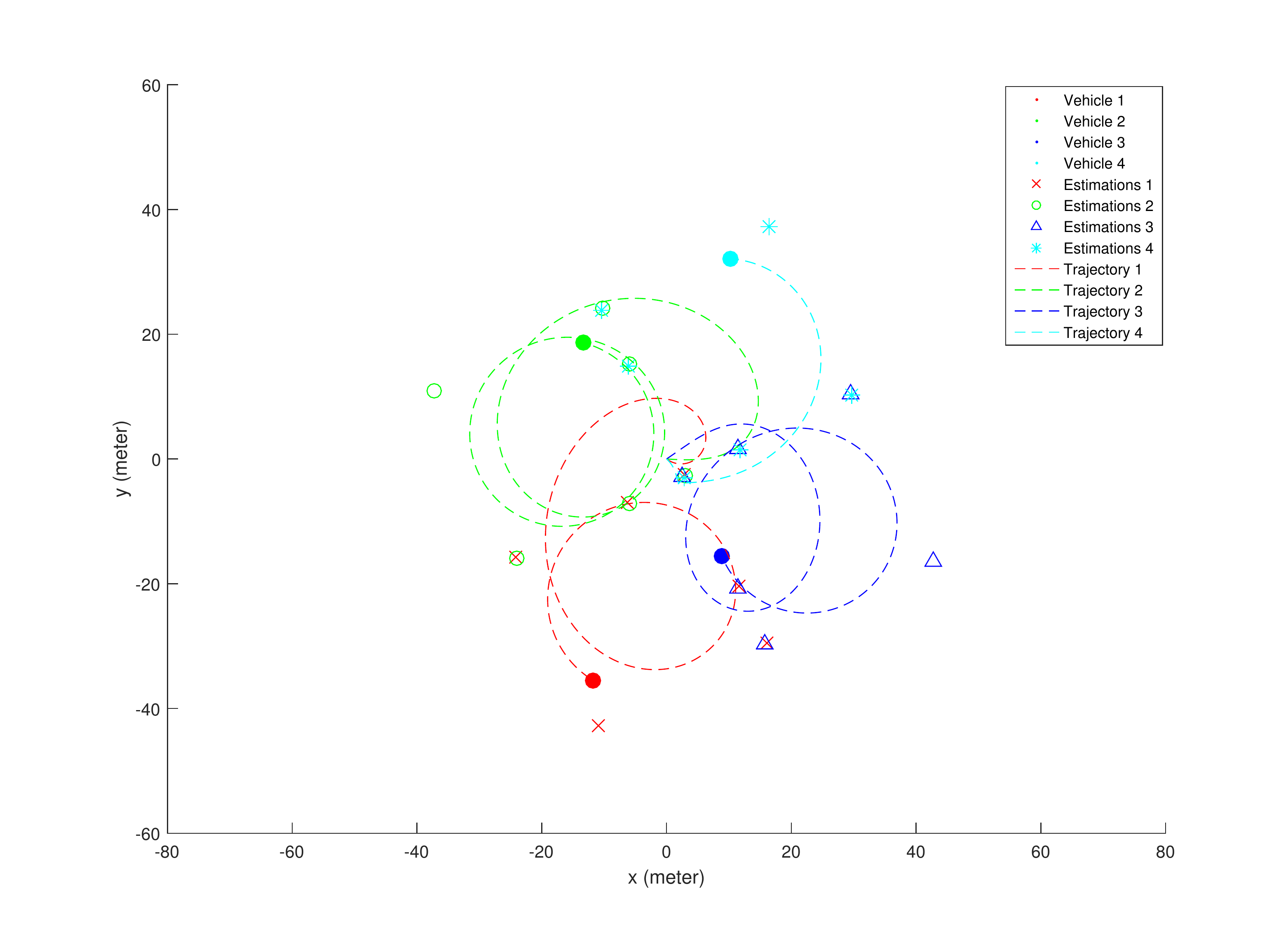}
\caption{Simulation results for cooperative SLAM with partial information. The red, green, blue and cyan landmarks and dashed lines of vehicle trajectories correspond respectively to estimations from vehicles 1, 2, 3 and 4}
\label{group2f}
\end{figure}

\subsection{Algorithm for collective localization with robots only}
There are cases that a swarm of robots need to localize each other and have a common map, also known as collective localization in literature. In these cases, there are no landmarks, and only robots in the map move around and localize each other and themselves. In such setting, a single robot $i$ will be able to measure relative heading difference between robot $i$ and robot $j$ as $\theta_{ij}$, most likely through a camera. In addition, it is normally assumed that all robots can observe all other robots in the swarm. Since in this problem there are no landmarks, we change the definition of ${\bf x}_{ik}$ to be the estimated location of robot $k$ in robot $i$'s coordinates. In this way, the problem of collective localization with only robots is very similar to the problem we presented in the first section about cooperative SLAM with full information. The only difference is that the previous landmarks are now moving vehicles. Since the vehicle is no different from the other vehicles it observes, we can use ${\bf x}_{ii}$ to denote robot $i$ in its own map instead of ${\bf x}_{iv}$. In that case, we can change what we have in Section 6.1 slightly into the same form, but with different variables as 
$$
\left[\begin{array}{c}\dot{\hat{{\bf x}}}_{ik}\\{\dot{\hat{{\bf x}}}_{ikv}}\end{array}\right]=({\bf v}_i+\Omega_i \left[\begin{array}{c}{\hat{{\bf x}}}_{ik}-{\hat{{\bf x}}}_{ic}\\{{\hat{{\bf x}}}_{ikv}}-{\hat{{\bf x}}}_{ic}\end{array}\right] )+ \left[\begin{array}{c}{\bf u}_{ij}\\{\bf u}_{ii}\end{array}\right]+{\bf P}_{ik}{\bf H}_{ik}^T{\bf R}^{-1}({\bf y}_{ik}-{\bf H}_{ik}\left[\begin{array}{c}{\hat{{\bf x}}}_{ik}\\{{\hat{{\bf x}}}_{ikv}}\end{array}\right])$$ 
$$
\dot{{\bf P}}_{ik}={\bf Q}_i-{\bf P}_{ik}{\bf H}_{ik}^T{\bf R}^{-1}{\bf H}_{ik}{\bf P}_{ik}$$
$$
{\bf x}_{iic}=(\sum_{k\in O}\Sigma_{ivk}^{-1})^{-1} \sum_{k\in O}(\Sigma_{ivk}^{-1}{\bf x}_{ivk})
$$
where  
$${\bf v}_i=\gamma_{vi}({\bf x}_{cc}-{\bf x}_{ic})$$
$$\omega_i=\gamma_{\omega_i}\Sigma_k({\bf x}_{ik}-{\bf x}_{ic})^T\begin{bmatrix}
0&1\\
-1&0
\end{bmatrix}{\bf x}_{ck}$$
and
$$\beta_{di}=argmin_{\beta_d\in[-\pi,\pi]}\Sigma_k({\bf y}_{ik}-{\bf H}_{ik}\left[\begin{array}{c}{\hat{{\bf x}}}_{ik}\\{{\hat{{\bf x}}}_{ikv}}\end{array}\right])^T({\bf y}_{ik}-{\bf H}_{ik}\left[\begin{array}{c}{\hat{{\bf x}}}_{ik}\\{{\hat{{\bf x}}}_{ikv}}\end{array}\right])$$
$$\dot{\hat{\beta}}_i=\omega_{im}+\omega_i+\gamma_{\beta_i}(\beta_{di}-\hat{\beta}_i)$$
$${\bf u}_{ii}=\begin{bmatrix}
u_i\sin\hat{\beta}_i\\
u_i\cos\hat{\beta}_i
\end{bmatrix}$$ 
$${\bf u}_{ij}=\begin{bmatrix}
u_j\sin(\hat{\beta}_i+\theta_{ij})\\
u_j\cos(\hat{\beta}_i+\theta_{ij})
\end{bmatrix}$$

\subsubsection{Simulation results}
The simulation environment for collective localization with robots only is slightly different from the ones we proposed before. Without any landmark, vehicles can observe each other and also measurements of relative headings. We use the same initial conditions for the simulation to implement algorithm for collective localization. 

As we show in the full video on \url{https://vimeo.com/193489767}, with no prior global information and only observation of other vehicles, coordinate systems from different vehicles shift and rotate to converge to each other. We can see that estimated positions of different vehicles converge to reach consensus, and trajectories of vehicles also converge to the circles they are expected to be. The achieved consensus result is a rotated and shifted transformation from the truth, which can be considered as the true map.

\begin{figure}[htbp] 
\centering 
\includegraphics[width=\textwidth]{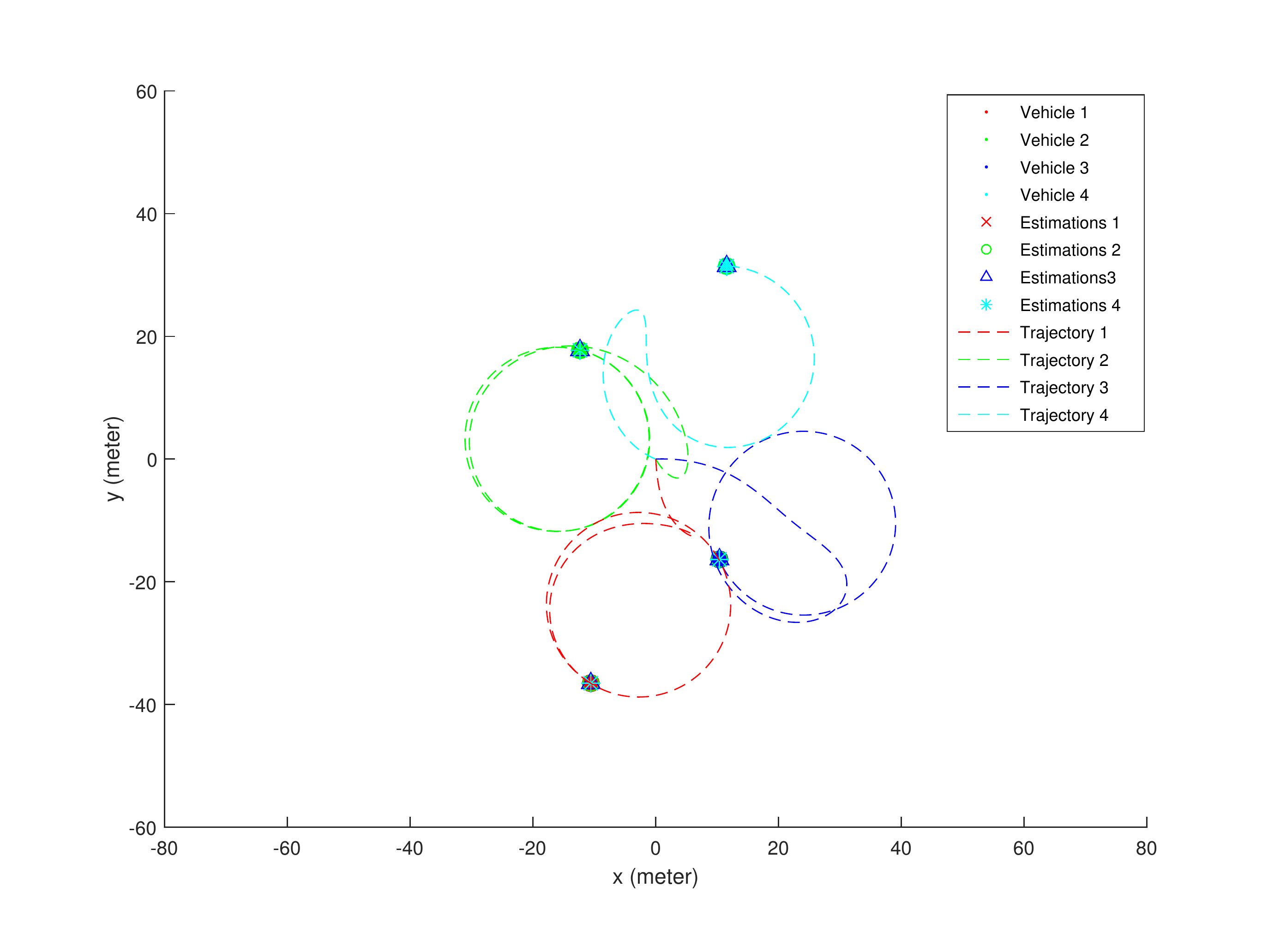}
\caption{Simulation results for collective localization with robots only. The red, green, blue and cyan dashed lines of vehicle trajectories correspond respectively to estimations from vehicles 1, 2, 3 and 4}
\label{group3f}
\end{figure}

\subsection{Remarks}
\subsubsection{Extension to 3D applications}
For the cases we discussed above, they are all in 2D settings. However, extending to 3D is very straightforward. For the translation part of the null space term
$${\bf v}_i=\gamma_{vi}({\bf x}_{cc}-{\bf x}_{ic})$$
it doesn't need to change from 2D to 3D. For the rotation part, since we are choosing the input $\Omega_i$ to minimize the heading error $e_h$ to zero. In the 2D case, we express the time derivative of heading error as a function of $\Omega_i$ as $\dot{e}_h(\omega_i)$ and choose $\omega_i$ to assure $$\dot{e}_h(\omega_i)\leq 0$$
So in the 3D case, it is nothing different. Since in 3D, the rotation part turns to 
$$\Omega_i=\begin{bmatrix}
0&-\omega_{iz}&\omega_{iy}\\
\omega_{iz}&0&-\omega_{ix}\\
-\omega_{iy}&\omega_{ix}&0
\end{bmatrix}$$
we can also use a vector ${\bf \omega}_i=[\omega_{ix},\omega_{iy}, \omega_{iz}]^T$ to model the time derivative of heading error as a function of ${\bf \omega}_i$ as $\dot{e}_h({\bf \omega}_i)$ and choose ${\bf \omega}_i$ to assure $$\dot{e}_h({\bf \omega}_i)\leq 0$$
In that case, extending our proposed algorithms to 3D is easy and straightforward.

\subsubsection{Extension to multi-camera pose estimation}

Small unmanned aerial vehicles (UAVs) have become popular robotic systems in recent years. Estimation of a small UAV's 6 degree of freedom (6 DOF) pose, relative to its surrounding environment using onboard cameras has also become more important. Results from the field of multi-camera egomotion estimation \cite{schauwecker2013board}\cite{kim2008motion}\cite{sola2008fusing}\cite{kaess2010probabilistic}\cite{ragab2010multiple}\cite{kim2007visual}\cite{baker2001spherical}\cite{pless2003using}\cite{harmat2015multi} show that such problem can be better solved by using multiple cameras positioned appropriately. When all cameras have been calibrated with precise positions and attitudes on the robot, it is straightforward to implement algorithms we proposed in Chapter 3 for multi-camera sensor fusion. As multiple cameras only add linear constraints to the LTV Kalman filter.

However, more frequently, it might be too complex or unrealistic to calibrate all cameras in advance. In such case, we can treat each single camera as a small ``robot'' with independent measurements. And the algorithm we proposed for multi-robot cooperative SLAM with partial information can be implemented on such applications, and poses of different cameras could be automatically calibrated, with one extra constraint that these cameras are fixed to one same robot and should have same translational and rotational velocities.

\section{Concluding Remarks}

In this paper, we propose using the combination of LTV Kalman filter and contraction tools to solve the problem of simultaneous mapping and localization (SLAM). By exploiting the virtual measurements, the LTV Kalman observer does not suffer from errors brought by the linearization process in the EKF SLAM, which makes the solution global and exact. Convergence rates can be quantified using contraction analysis.
The application cases utilize different kinds of sensor information that range from traditional bearing measurements and range measurements to novel ones like optical flows and time-to-contact measurements. They can solve SLAM problems in both 2D and 3D scenarios. Note that
\begin{itemize}

\item bounding of the covariance matrix ${\bf P}$ may be done analytically based
on the observability Grammian \citep{bryson1975applied, lohmiller2013contraction}. 

\item our approach is particularly suitable for exploiting the recent availability of vision sensors at very low cost, rather than relying on range sensors like lidars.

\item in the Victoria Park benchmark dataset, features are mostly trees in the park. As a result some regions have dense landmarks, while others  have sparse landmarks. Landmarks in dense areas and landmarks with high uncertainty provides less information for the updates on the states. Thus, incorporating feature selection to use landmarks with richer information could reduce computation workload like suggested in \cite{mu2013value} . Such active sensing could be achieved as in \citep{dickmanns1998vehicles, dickmanns2007dynamic, slotine2001modularity} by exploiting the fact that the posterior covariance matrix can be computed before taking any specific measurement. More generally, path planning may also be adjusted according to a desired exploration/exploitation trade-off \citep{vergassola2007infotaxis, schwager2009decentralized, mu2013value}.

\item it may be interesting to consider whether similar representations may  also be used in biological navigation, e.g. in the context of place cells or grid cells \citep{moser2008place} or sensing itself \citep{gollisch2010eye}.

\end{itemize}

\bibliographystyle{SageH}
\bibliography{science}
\end{document}